\documentclass[accepted,startpage=7314]{uai2026} %

\usepackage[american]{babel}

\usepackage[round,sort&compress]{natbib} %
\usepackage{mathtools} %
\usepackage[utf8]{inputenc} %
\usepackage[T1]{fontenc}    %
\usepackage{hyperref}       %
\usepackage{url}            %
\usepackage{booktabs}       %
\usepackage{amsfonts}       %
\usepackage{amsmath}        %
\usepackage{amssymb}        %
\usepackage{amsthm}         %
\usepackage{bbm}            %
\usepackage{enumitem}       %
\usepackage{wrapfig}        %
\usepackage{multirow}       %
\usepackage[skip=0pt]{caption}
\usepackage{nicefrac}       %
\usepackage{microtype}      %
\usepackage{xcolor}         %
\usepackage{pifont}         %
\usepackage{verbatim}       %
\usepackage{todonotes}      %
\usepackage{algorithm}      %
\usepackage{algpseudocodex} %
\usepackage{arydshln}
\usepackage{subcaption}
\usepackage[capitalise,noabbrev]{cleveref} %
\crefname{equation}{}{}

\theoremstyle{plain} %
\newtheorem{theorem}{Theorem}%
\newtheorem{lemma}{Lemma}%
\newtheorem{proposition}{Proposition}%
\newtheorem{corollary}{Corollary}%

\theoremstyle{definition} %
\newtheorem{assumption}{Assumption}%

\theoremstyle{remark} %
\newtheorem{remark}{Remark}%

\crefname{assumption}{Assumption}{Assumptions}

\DeclareMathOperator*{\argmin}{arg\,min}
\DeclareMathOperator*{\argmax}{arg\,max}
\DeclareMathOperator{\Tr}{Tr}
\DeclareMathOperator{\cxv}{cxv}
\DeclareMathOperator{\wrls}{WRLS}
\DeclareMathOperator{\wsb}{WSB}

\DeclareMathOperator{\diag}{diag}

\renewcommand{\P}{\mathbb{P}} %
\newcommand{\E}{\mathbb{E}} %
\newcommand{\R}{\mathbb{R}} %
\newcommand{\N}{\mathbb{N}} %
\newcommand{\I}{\mathbb{I}} %

\newcommand{\cX}{\mathcal{X}} %
\newcommand{\cF}{\mathcal{F}} %
\newcommand{\cE}{\mathcal{E}} %
\newcommand{\cN}{\mathcal{N}} %
\newcommand{\cD}{\mathcal{D}} %
\newcommand{\cH}{\mathcal{H}} %
\newcommand{\cO}{\mathcal{O}} %

\title{Weighted Sequential Bayesian Inference for Non-Stationary Linear Contextual Bandits}

\author[1]{\href{mailto:werge@sdu.dk}{Nicklas~Werge}}
\author[1,2]{\href{mailto:yishan.eason.wu@gmail.com}{Yi-Shan~Wu}}
\author[1]{\href{mailto:akgul@imada.sdu.dk}{Abdullah~Akgül}}
\author[1]{\href{mailto:kandemir@imada.sdu.dk}{Melih~Kandemir}}
\affil[1]{%
    Department of Mathematics and Computer Science\\ 
    University of Southern Denmark
}
\affil[2]{%
    Research Center for Information Technology Innovation\\ 
    Academia Sinica
}

\begin{document}
\maketitle

\begin{abstract}
In non-stationary linear contextual bandits, existing efficient algorithms typically rely on the Weighted Regularized Least-Squares (WRLS) estimator. Because WRLS only provides point estimates, previous methods typically construct surrogate distributions when aiming to perform Bayesian-like randomized exploration. To more properly establish the Bayesian principles, we introduce \emph{Weighted Sequential Bayesian} (WSB) inference, which forms a sequence of posteriors over a sequence of non-stationary reward parameters. This Bayesian take allows us to isolate the influence of initial beliefs into a dynamic prior penalty evaluated through the posterior covariance, which typically decreases over time. Building on this framework, we instantiate three WSB-based algorithms for exploration: \texttt{WSB-LinUCB}, \texttt{WSB-RandLinUCB}, and \texttt{WSB-LinTS}. By extending a refined drift analysis to randomized exploration without requiring local norms, we establish frequentist regret guarantees that match state-of-the-art WRLS-based baselines. Empirically, WSB's dynamic prior penalty reduces over-conservatism, allowing our algorithms to consistently match or exceed their WRLS-based counterparts. Lastly, we also provide a simplified proof for the time-uniform concentration of vector-valued martingales, a critical subroutine used throughout the literature, that might be of independent interest.
\end{abstract}

\section{Introduction} \label{sec::introduction}

Bandits provide a foundational framework for sequential decision-making under uncertainty, with applications spanning recommendation systems, clinical trials, and adaptive control~\citep{robbins1952some,bubeck2012regret,lattimore2020bandit}. Contextual bandits extend this framework by incorporating side information, like user characteristics, patient histories, or sensor readings, to better inform action selection~\citep{li2010contextual,krause2011contextual}. Among contextual bandits, their linear variant are particularly well-studied~\citep{chu2011contextual,dani2008stochastic,rusmevichientong2010linearly,abbasi2011improved}.

Many real-world applications involve \emph{non-stationarity}, in which the reward function evolves over time~\citep{besbes2014stochastic}. To cope with this, three main strategies have been proposed: \emph{restart}, which periodically reset the learning process~\citep{zhao2020simple}; \emph{sliding-window}, which focus on a fixed-size window of recent data~\citep{cheung2019learning}; and \emph{weighted}, which apply decaying weights to older data~\citep{russac2019weighted,wang2023revisit}.
Among these, weighted strategies are particularly appealing due to their smooth, continuous adaptation without requiring resets or fixed memory budgets. However, they have historically faced analytical challenges, and it remains an open question whether they can achieve optimal regrets~\citep{zhao2021non,touati2020efficient,faury2021technical,wang2023revisit}.

A fundamental challenge in bandit learning is the exploration-exploitation tradeoff. To balance this, exploration strategies generally fall into one of two paradigms: a \emph{frequentist} or a \emph{Bayesian} approach. The most common frequentist choices usually rely on the \emph{Weighted Regularized Least-Squares} (WRLS) estimator, which admits closed-form updates with per-round complexity $\mathcal{O}(d^2)$ where $d$ is the dimension, but such a point estimate lacks a native mechanism for uncertainty quantification. A Bayesian alternative can use, for instance, a \emph{Gaussian Process} (GP) objective \citep{srinivas2010gaussian,chowdhury2017kernelized,deng2022weighted}. While GPs provide calibrated uncertainty estimates in a \emph{non-parametric} framework, they require maintaining and inverting kernel matrices that scale with time $t$, resulting in a per-round complexity of $\mathcal{O}(t^3)$. This limits their use in long-horizon or streaming problems. There have been approaches trying to retain the $\mathcal{O}(d^2)$ computation complexity inspired by Thompson Sampling (TS) \citep{agrawal2013thompson,kim2020randomized,vaswani2020old}, which, however, did not take full advantage of the Bayesian power. In particular, the algorithms build surrogate posterior distributions from the WRLS point estimates, which are forced to inject artificial Gaussian noise scaled by frequentist covariance matrices. 

To bridge this gap, we revisit the non-stationary linear bandit problem from a parametric Bayesian perspective. Specifically, we introduce \emph{Weighted Sequential Bayesian} (WSB) inference, which copes with non-stationarity by building a sequence of generalized posteriors over the finite-dimensional, time-varying reward parameters $\theta_t \in \mathbb{R}^d$. While we show that the means of WSB posteriors algebraically coincide with the WRLS point estimates under an uninformative prior, our framework natively provides the posterior covariance required to properly ground randomized exploration. This allows WSB posteriors to retain the $\mathcal{O}(d^2)$ efficiency of WRLS estimators without relying on artificial surrogate distributions. Furthermore, by carefully incorporating this Bayesian structure into our concentration analysis, we show that the prior's influence shows up as a dynamic prior penalty term. In contrast to WRLS-based analyses, where the initialization penalty is bounded by a fixed worst-case constant, this WSB penalty is evaluated through the time-varying posterior covariance. This yields a data-dependent counterpart of the usual static prior penalty, which can be substantially smaller as the posterior covariance contracts.

\paragraph{Contributions.} 
The main contributions of this paper are as follows:
\begin{itemize}
    \item First, we establish a time-uniform confidence bound for WSB posteriors (\cref{lem::wsb::ucb}), which decomposes the estimation error into three distinct components: a drift term, a noise term, and a prior term. While this decomposition resembles standard techniques \citep{russac2019weighted,wang2023revisit}, our Bayesian approach allows us to explicitly isolate the influence of the initial beliefs into a dynamic prior penalty term ($\Pi_t$).
    \item Second, while the self-normalized bound used to control the noise term is a weighted corollary to classical martingale results \citep{abbasi2011improved}, we provide a simplified proof of this vector-valued concentration inequality. Inspired by the advances in modern martingale theory \citep{howard2020time}, we directly apply Ville's inequality for non-negative super-martingales \citep{doob1939-ville}, bypassing the complex stopping-time constructions traditionally used in the bandit literature.
    \item Third, building on the WSB framework, we propose three exploration strategies: Upper Confidence Bound (UCB), randomized UCB (RandUCB), and Thompson Sampling (TS), yielding \texttt{WSB-LinUCB}, \texttt{WSB-RandLinUCB}, and \texttt{WSB-LinTS}, respectively. We establish frequentist regret guarantees for all three algorithms, summarized in \cref{tab::related_work}. While the $d^{1/8}$ regret improvement over earlier baselines relies on the refined drift analysis of \citet{wang2023revisit}, which had previously been applied only to UCB-based approaches, we extend these guarantees to randomized exploration strategies. Furthermore, our experiments demonstrate the empirical benefits of this Bayesian treatment across different dimensions. Since WSB evaluates the prior penalty through the time-varying posterior covariance rather than a fixed worst-case constant, its arm-selection criteria tends to be less conservative.
\end{itemize}

\begin{table}[h]
\caption{Regrets for non-stationary linear contextual bandits; $d$ is the dimension, $K$ the number of actions, $B_T$ the non-stationarity, $T$ the horizon.}
\label{tab::related_work}
\begin{center}
\small
\resizebox{\columnwidth}{!}{%
\begin{tabular}{lc}
\toprule
\textbf{ALGORITHM} & \textbf{REGRET} \\ 
\midrule
\texttt{D-LinUCB} \citep{russac2019weighted} & $\tilde{\mathcal{O}}(d^{7/8}B_{T}^{1/4}T^{3/4})$ \\
\texttt{LB-WeightUCB} \citep{wang2023revisit} & $\tilde{\mathcal{O}}(d^{3/4}B_{T}^{1/4}T^{3/4})$ \\
\texttt{WSB-LinUCB} (\textbf{Ours}, \cref{alg:wsb_linucb}) & $\tilde{\mathcal{O}}(d^{3/4}B_{T}^{1/4}T^{3/4})$ \\
\midrule
\texttt{D-RandLinUCB} \citep{kim2020randomized} & $\tilde{\mathcal{O}}(d^{7/8}B_{T}^{1/4}T^{3/4})$ \\ 
\texttt{WSB-RandLinUCB} (\textbf{Ours}, \cref{alg:wsb_randlinucb}) & $\tilde{\mathcal{O}}(d^{3/4}B_{T}^{1/4}T^{3/4})$ \\
\midrule
\texttt{D-LinTS} \citep{kim2020randomized} & $\tilde{\mathcal{O}}(d^{7/8}\log(K)^{3/8}B_{T}^{1/4}T^{3/4})$ \\
\texttt{WSB-LinTS} (\textbf{Ours}, \cref{alg:wsb_lints}) & $\tilde{\mathcal{O}}(d^{3/4}\log(K)^{3/8}B_{T}^{1/4}T^{3/4})$ \\ 
\bottomrule
\end{tabular}}
\end{center}
\end{table}

\paragraph{Notations.}
For $x,y \in \R^{d}$, let $\langle x, y \rangle$ denote the standard inner product and $\lVert x \rVert_{2} = \sqrt{\langle x, x \rangle}$ the Euclidean norm. Given a matrix $M \in \R^{d \times d}$, we define the weighted inner product $\langle x, y \rangle_{M}=x^\top M y$ and weighted norm $\lVert x \rVert_{M} = \sqrt{x^\top M x}$. Let $\lambda_{\min}(M)$ and $\lambda_{\max}(M)$ denote the smallest and largest eigenvalues of $M$, respectively. We write $M \succ 0$ for positive definite and $M \succeq 0$ for positive semi-definite matrices. Let $\mathbb{I}_{d}$ be the $d \times d$ identity matrix.

\section{Background} \label{sec::background}

Before our Bayesian treatment, we first present the non-stationary linear contextual bandit problem. We then review the widely used \emph{Weighted Regularized Least-Squares} (WRLS) estimator, which forms the basis of many existing algorithms. In particular, we discuss how WRLS is used in both (deterministic) UCB-based and randomized exploration strategies, and highlight recent analytical refinements that improve regrets.

\subsection{Problem Formulation} \label{sec::background::problem}
The non-stationary linear contextual bandit problem is defined as follows. At each round $t$, the learner observes a set of $K$ context-dependent actions $\cX_{t} \subseteq \R^{d}$, which may change. Based on the history from the previous $t-1$ rounds, denoted by $\cH_{t-1} = \{(X_s, r_s)\}_{s=1}^{t-1}$, the learner selects an action $X_t \in \cX_{t}$ and receives a noisy reward $r_{t} = \langle X_{t}, \theta_{t}^{*} \rangle + \varepsilon_{t}$, where $\theta_{t}^{*} \in \R^{d}$ is the unknown time-varying reward parameter and $\varepsilon_{t}$ is conditionally $\sigma$-sub-Gaussian given the $\sigma$-algebra $\mathcal{F}_{t-1} = \sigma(\cH_{t-1}, X_t)$; that is, for all $\nu \in \R$, $\E[\exp(\nu \varepsilon_{t}) \vert \cF_{t-1}] \leq \exp( \nu^{2} \sigma^{2}/2)$ almost surely. We make the following standard boundedness assumption throughout:
\begin{assumption} \label{ass:bounded_parameters_features}
There exist $S,L\geq0$ such that for all $t$, $\lVert \theta_{t}^{*} \rVert_{2} \leq S$ and $\lVert x \rVert_{2} \leq L$ for any  $x \in \cX_{t}$.
\end{assumption}
The degree of non-stationarity is measured by the total variation budget $B_{T} = \sum_{t=1}^{T-1} \lVert \theta_{t}^{*}-\theta_{t+1}^{*} \rVert_{2}$, which accounts for both gradual drifts and abrupt changes in the underlying reward parameters \citep{besbes2014stochastic,garivier2011upper}. Let $X_{t}^{*}=\argmax_{x\in\cX_t}\langle x ,\theta_{t}^{*}\rangle$ denote the best action in round $t$. The learner's goal is to minimize the dynamic (pseudo-)regret 
$$R_{T} = \sum_{t=1}^{T} \langle X_{t}^{*}, \theta_{t}^{*} \rangle - \sum_{t=1}^{T} \langle X_{t}, \theta_{t}^{*} \rangle.$$

\subsection{Weighted Regularized Least-Squares} \label{sec::background::wrls}

In this section, we review the WRLS estimator, its role in UCB-based exploration under the refined analysis of \citet{wang2023revisit}, and the construction of randomized exploration strategies upon it. Later, in \cref{sec::wsb}, this WRLS estimator will be replaced by its Bayesian analogue, which retains a similar recursive form while incorporating prior beliefs.

\paragraph{WRLS estimator.}
The WRLS estimator is a popular choice in non-stationary settings because it adapts to evolving parameters by down-weighting past observations through exponential discounting \citep{russac2019weighted,kim2020randomized,wang2023revisit}. It is defined as:
\begin{equation} \label{eq::wrls}
    \hat{\theta}_{t} = \argmin_{\theta\in\R^d} \left(\lambda\lVert \theta \rVert_{2}^{2} + \sum_{s=1}^{t} \gamma^{t-s}( \langle X_{s}, \theta \rangle - r_{s})^{2} \right),
\end{equation}
where $\lambda>0$ is a regularization parameter and $\gamma\in(0,1)$ is a discounted factor. If $\gamma = 1$, this reduces to the Regularized Least-Squares (RLS) estimator, which is widely used in stationary settings \citep{abbasi2011improved,agrawal2013thompson,abeille2017linear,vaswani2020old}.

The WRLS estimator admits a closed-form solution given by $\hat{\theta}_{t} = V_{t}^{-1} b_{t}$, where $V_{t} = \lambda \I_{d} + \sum_{s=1}^{t} \gamma^{t-s} X_{s} X_{s}^{\top}$ and $b_{t} = \sum_{s=1}^{t} \gamma^{t-s} X_{s} r_{s}$. The matrix $V_{t}\in\R^{d  \times d}$ is positive definite by construction, and both $V_{t}$ and $b_{t}$ can be updated recursively; $V_{t} = \gamma V_{t-1} + X_{t} X_{t}^\top + (1 - \gamma) \lambda \mathbb{I}_d$ and $b_t = \gamma b_{t-1} + X_{t} r_{t}$, with initial values $V_1 = \lambda \I_{d}$ and $b_1 = \mathbf{0}$. Consequently, the WRLS estimator can be updated recursively as $\hat{\theta}_{t} = V_{t}^{-1} b_{t}$, initialized with $\hat{\theta}_1 = \mathbf{0}$, and thereby avoids the need to store past observations.

\paragraph{Error decomposition.}
Existing analyses of WRLS-based algorithms typically decompose the estimation error $(\hat{\theta}_{t-1} - \theta_{t}^{*})$ into two distinct parts: a \emph{drift part}, capturing non-stationarity in the reward parameters, and a \emph{noise part}, reflecting randomness in the observed rewards. The index shift arises because the decision at round $t$ is based on $\hat{\theta}_{t-1}$, computed from observations available only up to round $t-1$. The \emph{drift part} is bounded by the bias term $\alpha_{t}^{\wrls}$, while the \emph{noise part} is captured by the confidence radius $\beta_{t}^{\wrls}(\delta)$.

Earlier work, such as \citet{cheung2022hedging,russac2019weighted,kim2020randomized}, controlled the \emph{drift part} using sliding-window techniques \citep{cheung2019learning}, which retain only a fixed-length subset of recent observations. While intuitive, this approach introduces unnecessary complexity. In contrast, \citet[Lemma~4]{wang2023revisit} proposed a simpler analysis that yields a fully deterministic bound on the drift without relying on artificial windows.

For the \emph{noise part}, high-probability concentration inequalities are used. In the stationary setting, \citet{abbasi2011improved} derived a self-normalized Martingale tail bound for the RLS estimator, later generalized to non-stationary problems by \citet{russac2019weighted} via the \emph{local norm}, a construct adopted in several subsequent studies \citep{touati2020efficient,kim2020randomized}. However, this \emph{local norm} is merely a technical artifact of the analysis rather than a fundamental necessity; see, e.g., \citet[Lemma~5]{wang2023revisit}.

\paragraph{Concentration bounds for WRLS.}
The bounds on these two components, drift and noise, together control the estimation error $(\hat{\theta}_{t-1} - \theta_{t}^{*})$. These bounds are central for constructing exploration strategies.
\begin{lemma}[{\citet[Lemma~1]{wang2023revisit}}] \label{lem::wrls::ucb}
For any $\delta \in (0,1)$, with probability at least $1-\delta$, the following inequality holds for all $t \in \mathbb{N}_{+}$:
\begin{equation*}
    \lVert \hat{\theta}_{t-1} - \theta_{t}^{*} \rVert_{V_{t-1}} \leq \alpha_{t-1}^{\wrls} + \beta_{t-1}^{\wrls}(\delta)
\end{equation*}
and $\forall x \in \cX_{t}, \; \lvert \langle x , \hat{\theta}_{t-1} - \theta_{t}^{*} \rangle \rvert \leq (\alpha_{t-1}^{\wrls} + \beta_{t-1}^{\wrls}(\delta)) \lVert x \rVert_{V_{t-1}^{-1}}$, where 
\begin{itemize}
    \item $\alpha_{t}^{\wrls} = L \sqrt{d} \sum_{k=1}^{t} \sqrt{\gamma^{t}} \sqrt{\frac{\gamma^{-k}-1}{1-\gamma}} \lVert \theta_{k}^{*} - \theta_{k+1}^{*} \rVert_{2}$ is the drift-induced bias and 
    \item $\beta_{t}^{\wrls}(\delta) = \sigma\sqrt{2\log(\frac{1}{\delta})+d\log(1+\frac{L^2(1-\gamma^{2t})}{\lambda d(1-\gamma^2)})} +  \sqrt{\lambda}S$ is the confidence level.
\end{itemize}
\end{lemma}

\Cref{lem::wrls::ucb} can be further simplified using $\lVert x \rVert_{2} \leq L$ (\cref{ass:bounded_parameters_features}), and the property that for a positive definite and symmetric matrix $M\in\R^{d\times d}$ and any vector $x\in\R^d$, $\lVert x \rVert_{M^{-1}} \leq \lVert x \rVert_{2}/\sqrt{\lambda_{\min}(M)}$, yielding $\alpha_{t-1}^{\wrls} \lVert x \rVert_{V_{t-1}^{-1}} \leq L \alpha_{t-1}^{\wrls} / \sqrt{\lambda}$ that is independent of $x$.

\subsection{Deterministic Exploration with WRLS} \label{sec::background::ucb}
\citet{wang2023revisit} use \cref{lem::wrls::ucb} to form their UCB-based algorithm, \texttt{LB-WeightUCB}. Specifically, they take
\begin{equation*}
X_{t} = \argmax_{x \in \cX_{t}} \{ \langle x, \hat{\theta}_{t-1} \rangle + \beta_{t-1}^{\wrls}(\delta) \lVert x \rVert_{V_{t-1}^{-1}} \}.
\end{equation*}
Notably, \texttt{LB-WeightUCB} requires maintaining only a single covariance matrix $V_{t}$. An earlier UCB-based algorithm, \texttt{D-LinUCB} by \citet{russac2019weighted}, constructs upper confidence bounds using a more complex \emph{local norm}. Specifically, they set 
\begin{equation*}
X_{t} = \argmax_{x \in \cX_{t}} \{ \langle x, \hat{\theta}_{t-1} \rangle + \beta_{t-1}^{\wrls}(\delta) \lVert x \rVert_{V_{t-1}^{-1} \tilde{V}_{t-1} V_{t-1}^{-1}} \},
\end{equation*}
where $\tilde{V}_{t} = \lambda \I_{d} + \sum_{s=1}^{t} \gamma^{2(t-s)} X_s X_s^\top$ is an additional covariance matrix in $\mathbb{R}^{d \times d}$. Obviously, \texttt{D-LinUCB} is both computationally and memory-wise more demanding than \texttt{LB-WeightUCB}, since it requires maintaining and inverting two $d \times d$ matrices, whereas \texttt{LB-WeightUCB} works with just one. For instance, \citet[Figure~1]{wang2023revisit} report that \texttt{LB-WeightUCB} achieves over a $1.5\times$ speedup relative to \texttt{D-LinUCB}, mainly due to eliminating the \emph{local norm} and relying solely on $V_{t}$.

\paragraph{Regret guarantees.}
The established confidence bound is central to the analysis of UCB-style algorithms that use WRLS. As summarized in \cref{tab::related_work}, \texttt{LB-WeightUCB}~\citep{wang2023revisit} achieves a regret of $\tilde{\mathcal{O}}(d^{3/4} B_{T}^{1/4} T^{3/4})$, while the earlier \texttt{D-LinUCB} \citep{russac2019weighted} has a regret of $\tilde{\mathcal{O}}(d^{7/8} B_{T}^{1/4} T^{3/4})$. This improved regret is a direct consequence of the refined treatment described above.

\subsection{Randomized Exploration with WRLS} \label{sec::background::randexp}
We now review two randomized exploration strategies based on the WRLS estimator: \texttt{D-RandLinUCB} and \texttt{D-LinTS} \citep{kim2020randomized}. Like \texttt{D-LinUCB}, both build on WRLS estimates and use \emph{local norms} in their confidence bounds, which require maintaining two covariance matrices (see \cref{sec::background::ucb}). Unlike deterministic UCB-based algorithms that address uncertainty via the Optimism-in-the-Face-of-Uncertainty (OFU) principle, the randomized approaches introduce randomness into the action selection process: \texttt{D-RandLinUCB} perturbs the confidence level, while \texttt{D-LinTS} samples randomized parameter estimates. We describe each method in detail below.

\paragraph{Randomized UCB.}
Randomized UCB was originally proposed by \citet{vaswani2020old} and later extended to the non-stationary setting by \citet{kim2020randomized}. In this approach, the confidence level $\beta_{t}^{\wrls}(\delta)$ in \texttt{D-LinUCB} is replaced by a random variable $\eta_{t}$, sampled from a fixed, easy-to-sample distribution with confidence level $a>0$. For example, \citet[Section 4]{kim2020randomized} use a truncated univariate Gaussian distribution that assigns probability mass only to $[0, \infty)$ in their numerical experiments. At each round $t$, a sample $\eta_{t} \sim \cN(0,a^{2})$ is drawn,\footnote{For simplicity we present the Gaussian case. More generally, one may sample from any distribution $\cD(\delta,a)$ that satisfies suitable concentration and anti-concentration properties \citep{kim2020randomized,vaswani2020old,kveton2020perturbed}.} and the action is selected according to
\begin{equation*}
X_{t} = \argmax_{x \in \cX_{t}} \{ \langle x, \hat{\theta}_{t-1} \rangle + \eta_{t} \lVert x \rVert_{V_{t-1}^{-1} \tilde{V}_{t-1} V_{t-1}^{-1}} \}.
\end{equation*}

\paragraph{Thompson Sampling.}
The idea of injecting noise into the arm-selection criteria is not new. One of the most well-known ways of perturbing the estimates is Thompson Sampling (TS) \citep{thompson1933likelihood}. For stationary linear contextual bandits, TS is known as \texttt{LinTS}~\citep{agrawal2013thompson}, while \texttt{D-LinTS}~\citep{kim2020randomized} is its non-stationary counterpart.
The \texttt{D-LinTS} algorithm follows the recipe of \citet{abeille2017linear} that builds a sequence of posteriors with similar components as UCB-based methods. In particular, at each round $t$, a randomized parameter is sampled as $\tilde{\theta}_{t-1} = \hat{\theta}_{t-1} + V_{t-1}^{-1} \tilde{V}_{t-1}^{1/2} \eta_{t}$, where $\eta_{t}\sim\cN(0,a^{2}\I_{d})$. The action is then selected according to
\begin{equation*}
X_{t} = \argmax_{x \in \cX_{t}} \{ \langle x, \tilde{\theta}_{t-1} \rangle\}.
\end{equation*}

\paragraph{Coupled and decoupled randomization.}
A key distinction between \texttt{D-RandLinUCB} and \texttt{D-LinTS} lies in how random perturbations are applied when selecting actions. The arm-selection criterion of \texttt{D-LinTS} can be written as
\begin{equation*}
\langle x, \tilde{\theta}_{t-1} \rangle = \langle x, \hat{\theta}_{t-1} \rangle + x^{\top} V_{t-1}^{-1} \tilde V_{t-1}^{1/2}\eta_{t},
\end{equation*}
which equivalently can be expressed as
\begin{equation*}
\langle x, \hat{\theta}_{t-1} \rangle + \eta_{t,x}\,\|x\|_{V_{t-1}^{-1}\tilde V_{t-1}V_{t-1}^{-1}}, \quad \eta_{t,x}\sim\cN(0,a^{2}).
\end{equation*}
In \texttt{D-RandLinUCB}, the random perturbation is \emph{coupled}, i.e., the same random variable is used for all arms in a given round. This means that the randomness affects all actions simultaneously, resulting in correlated exploration across the action set. In contrast, \texttt{D-LinTS} uses \emph{decoupled} perturbations, where each arm receives an independent random perturbation in every round. This leads to greater variability in the exploration process, but also results in a slightly higher regret bound due to the increased variance.

\paragraph{Regret guarantees.}
As outlined in \cref{tab::related_work}, the regret of \texttt{D-RandLinUCB} is $\tilde{\mathcal{O}}(d^{7/8} B_{T}^{1/4} T^{3/4})$ and \texttt{D-LinTS} is $\tilde{\mathcal{O}}(d^{7/8}\log(K)^{3/8}B_{T}^{1/4} T^{3/4})$. Notably, if the refined analysis of \citet{wang2023revisit} is applied in place of that of \citet{russac2019weighted} as described in \cref{sec::background::ucb}, these regrets can be improved to $\tilde{\mathcal{O}}(d^{3/4} B_{T}^{1/4} T^{3/4})$ for \texttt{D-RandLinUCB} and $\tilde{\mathcal{O}}(d^{3/4}\log(K)^{3/8} B_{T}^{1/4} T^{3/4})$ for \texttt{D-LinTS}.

\section{A Bayesian Treatment} \label{sec::wsb}

While randomized algorithms such as \texttt{D-RandLinUCB} and \texttt{D-LinTS} introduce probabilistic exploration, they remain fundamentally rooted in frequentist point estimation. Because the WRLS objective only outputs point estimates $\hat{\theta}_t$, these algorithms construct a surrogate distribution by injecting artificial Gaussian noise scaled by the matrix $V_t$. 
This motivates us to develop a more principled tool that properly grounds these randomized algorithms.

In this section, we introduce Weighted Sequential Bayesian (WSB) inference that is able to handle the time-varying reward parameter $\theta_t$. By maintaining the WSB posteriors, we establish the foundation of randomized exploration, and as a natural byproduct, we obtain a dynamic prior penalty ($\Pi_t$) in our concentration bounds that frequentist point estimates do not capture.

While previous work usually focuses on exponential weighting, we consider general weight sequences $\{w_{s,t} \in [0,1] : 1 \leq s \leq t\}$ that are non-decreasing in $s$ (i.e., $w_{s-1,t} \leq w_{s,t}$) and satisfy the multiplicative consistency condition $w_{s,t} \geq w_{t-1,t}  w_{s,t-1}$. This includes exponential weights $w_{s,t} = \gamma^{t-s}$ for some $\gamma \in (0,1)$, which we adopt as a canonical example throughout the paper.

\paragraph{WSB posteriors.}
For analytical tractability, we assume Gaussian reward noise with known variance $\sigma^{2}>0$; this is not a restriction since prior algorithms also use $\sigma$ when defining UCB confidence levels (see, e.g., \cref{lem::wrls::ucb}). For any Gaussian prior $\pi(\theta) = \cN(\theta \vert \mu_0, \Sigma_0)$ with mean $\mu_{0} \in \R^{d}$ and covariance $\Sigma_{0} \in \R^{d \times d}$, the posterior distribution over $\theta$ at round $t$ remains Gaussian. We denote this posterior by $\rho_t(\theta) = \cN(\theta \vert \mu_t, \Sigma_t)$, where the posterior mean and covariance are given as follows:
\begin{equation} \label{eq::posterior::mean}
\mu_{t} = \Sigma_{t}\left(\Sigma_{0}^{-1}\mu_{0}+\frac{1}{\sigma^{2}}\sum_{s=1}^{t}w_{s,t}X_{s}r_{s} \right)
\end{equation}
and
\begin{equation} \label{eq::posterior::cov}
\Sigma_{t}^{-1} = \Sigma_{0}^{-1}+\frac{1}{\sigma^{2}}\sum_{s=1}^{t}w_{s,t}X_{s}X_{s}^{\top}.
\end{equation}
With exponential weighting, defined by $w_{s,t} = \gamma^{t-s}$ for some $\gamma \in (0,1)$, the updates in \cref{eq::posterior::mean,eq::posterior::cov} simplify into recursive forms, requiring no storage of past data: 
\begin{align} 
\mu_t &= \Sigma_t\left(\gamma \Sigma_{t-1}^{-1}\mu_{t-1} + \sigma^{-2}X_t r_t + (1-\gamma)\Sigma_0^{-1}\mu_0\right) \nonumber
\\ \Sigma_t^{-1} &= \gamma \Sigma_{t-1}^{-1} + \sigma^{-2}X_t X_t^\top + (1-\gamma)\Sigma_0^{-1}. \label{eq:WSB-posterior-update}
\end{align}

\begin{remark}
    The WSB posterior corresponds to applying Bayes' rule with tempered likelihoods $\rho_t(\theta) \propto \rho_0(\theta) \prod_{s=1}^t p(r_s|X_s,\theta)^{w_{s,t}}$, which defines a generalized Bayesian posterior~\citep{bissiri2016general,grunwald2017inconsistency}. This mirrors the design in non-parametric approaches such as weighted Gaussian Processes~\citep{deng2022weighted}, which is necessary to handle non-stationarity.
\end{remark}

\begin{remark}
    Under an uninformative prior ($\mu_0 = \mathbf{0}$, $\Sigma_0^{-1} = \lambda \sigma^2 \mathbb{I}_d$), the WSB posterior mean $\mu_t$ recovers the WRLS estimate $\hat{\theta}_t$. Thus, WSB naturally generalizes WRLS-based approaches; while matching their point estimates, it natively provides a full posterior covariance for sampling. Furthermore, while the WRLS initialization penalty ($\lambda \|\theta\|_2^2$) is bounded by a static constant $\sqrt{\lambda}S$, WSB evaluates its prior penalty ($\|\theta - \mu_0\|_{\Sigma_0^{-1}}^2$) through the shrinking posterior covariance $\Sigma_t$. This yields a dynamic prior penalty term ($\Pi_t$) in our concentration bounds that tightens over time.
\end{remark}
\paragraph{Concentration bounds for WSB posteriors.}
We also provide concentration bounds for the WSB posteriors. Our analysis mirrors the improved decomposition for WRLS from \citet{wang2023revisit} by separating parameter drift from stochastic noise. A key distinction, however, is that we isolate the prior penalty $\Pi_t$ from the noise term $\beta_t^{\wsb}$, allowing it to be controlled separately in~\cref{lem::wsb::prior_term}.
\begin{lemma} \label{lem::wsb::ucb}
For any horizon $T \in \N_{+}$, error probability $\delta \in (0,1)$, and prior $\pi(\theta) = \mathcal{N}(\theta \vert \mu_{0},\Sigma_{0})$, with probability at least $1-\delta$, the following inequalities hold for all posteriors $\rho_{t}(\theta) = \cN(\theta \vert \mu_{t},\Sigma_{t})$ for all $t \in [T]$ simultaneously:
\begin{equation*}
    \lVert\mu_{t-1}-\theta_{t}^{*}\rVert_{\Sigma_{t-1}^{-1}} \leq \alpha_{t-1}^{\wsb} + \beta_{t-1}^{\wsb}(\delta/T) + \Pi_{t-1}
\end{equation*}
and $\forall x \in \cX_{t}, \; \lvert \langle x, \mu_{t-1}-\theta_{t}^{*} \rangle \rvert \leq (\alpha_{t-1}^{\wsb} + \beta_{t-1}^{\wsb}(\delta/T) + \Pi_{t-1}) \lVert x \rVert_{\Sigma_{t-1}}$, where
\begin{itemize}
    \item $\alpha_{t}^{\wsb} = L \sqrt{d} \sum_{k=1}^{t} \sqrt{\sum_{s=1}^{k} w_{s,t}} \lVert \theta_{k}^{*}-\theta_{k+1}^{*} \rVert_{2}$ is the drift-induced bias,
    \item $\beta_{t}^{\wsb}(\delta') = \sqrt{2\log (\frac{1}{\delta'})+d\log(1+\frac{\Tr(\Sigma_{0})L^{2}\sum_{s=1}^{t}w_{s,t}^{2}}{d\sigma^{2}})}$ is the confidence level evaluated at a given failure allocation $\delta'$, and 
    \item $\Pi_{t}=\max_{\theta:\lVert\theta\rVert_{2}\leq S}\rVert\Sigma_{0}^{-1}(\mu_{0}-\theta)\rVert_{\Sigma_{t}}$ is the time-decaying prior term.
\end{itemize}
\end{lemma}

The proof of \cref{lem::wsb::ucb} is given in \cref{sec::appendix:::wbs}. The projection bound for arbitrary actions $x$ follows by the dual norm inequality $|\langle x,z\rangle| \leq \|x\|_{\Sigma_{t-1}}\|z\|_{\Sigma_{t-1}^{-1}}$, making this step explicit. The drift term $\alpha_{t}^{\wsb}$, with $w_{s,t} = \gamma^{t-s}$, essentially coincides with $\alpha_t^{\wrls}$ in WRLS that accumulates the effects of non-stationarity. 
The confidence level $\beta_{t}^{\wsb}$ captures the stochastic noise only. Unlike $\beta_{t}^{\wrls}$, which also absorbs the regularization term $\sqrt{\lambda}S$, our formulation isolates this regularization effect to the prior term $\Pi_t$.

Calculating $\Pi_{t}$ exactly requires solving a maximization problem, which is computationally impractical. To address this, we provide two tractable upper bounds on $\Pi_{t}$ that will be useful in both analysis and implementation.
\begin{lemma} \label{lem::wsb::prior_term}
For any prior $\pi(\theta) = \mathcal{N}(\theta \vert \mu_{0},\Sigma_{0})$, the following inequalities hold for all posteriors $\rho_{t}(\theta) = \mathcal{N}(\theta \vert \mu_{t},\Sigma_{t})$ and $t \in [T]$, simultaneously:
\begin{equation*}
\Pi_{t} \leq \Pi_{t}^{\cxv} \leq \Pi_{t}^{\Delta},
\end{equation*}
where
\begin{itemize}
    \item $(\Pi_{t}^{\cxv} )^{2} = \lVert \mu_{0} \rVert_{M_{t}}^{2} + S^2 \lambda_{\max}(M_t) + 2 S \sqrt{\mu_0^\top M_t^2 \mu_0}$
    \item $\Pi_{t}^{\Delta} = \lVert \mu_{0} \rVert_{M_t} + \sqrt{\lambda_{\max}(M_{t})}S$
\end{itemize}
with $M_t = \Sigma_{0}^{-1}\Sigma_{t}\Sigma_{0}^{-1}$. 
\end{lemma}

The proof of \cref{lem::wsb::prior_term} can be found in \cref{sec::appendix:::wbs}. The bound $\Pi_{t}^{\cxv}$ handles the interaction across the full spectrum of $M_t$ to resolve alignment issues, while the simpler upper bound $\Pi_{t}^{\Delta}$ remains easier to evaluate and suffices for most theoretical purposes. To explicitly see how this penalty dynamically improves upon the static WRLS bound, we note that as $\Sigma_t$ contracts, $\Pi_t\to0$, unlike the fixed $\sqrt{\lambda}S$ term.

\subsection{Proof Sketch of Lemma~\ref{lem::wsb::ucb}}
\Cref{lem::wsb::ucb} provides a unified bound on the \emph{estimation error} around the WSB posterior mean $\mu_{t-1}$ and the current reward parameter $\theta_{t}^{*}$. To establish this bound, we follow a common approach in the literature (see e.g., \citet{abbasi2011improved,russac2019weighted,wang2023revisit}), using a \emph{noise-free surrogate} to separate the contributions of parameter drift and stochastic noise. Specifically, we define the \emph{noise-free surrogate posterior mean}:
\begin{equation} \label{eq::posterior::surrogate::mean}
\bar{\mu}_{t} = \Sigma_{t-1}\left(\Sigma_{0}^{-1}\theta_{t}^{*}+\frac{1}{\sigma^{2}}\sum_{s=1}^{t-1}w_{s,t-1}X_{s}X_{s}^{\top}\theta_{s}^{*} \right),
\end{equation}
which allows us to decompose the \emph{estimation error} $(\mu_{t-1} - \theta_{t}^{*})$ into two terms: a \emph{drift part} $(\bar{\mu}_{t} - \theta_{t}^{*})$ and a \emph{noise part} $(\mu_{t-1} - \bar{\mu}_{t})$, which we can control independently.

The \emph{drift part} captures the deviation caused by non-stationarity. As this part is unaffected by noise, it can be bounded deterministically, mirroring the bias control in WRLS, but with general weighting $w_{s,t}$:
\begin{lemma} \label{lem::wsb::drift}
Given the prior $\pi(\theta) = \mathcal{N}(\theta \vert \mu_{0},\Sigma_{0})$, the following inequalities hold for all posteriors $\rho_{t}(\theta) = \mathcal{N}(\theta \vert \mu_{t},\Sigma_{t})$ and $t \in \N_+$:
\begin{equation*}
    \lVert\bar{\mu}_{t}-\theta_{t}^{*}\rVert_{\Sigma_{t-1}^{-1}} \leq \alpha_{t-1}^{\wsb}
\end{equation*}
and $\forall x \in \cX_{t}, \; \lvert \langle x, \bar{\mu}_{t}-\theta_{t}^{*} \rangle \rvert \leq \alpha_{t-1}^{\wsb} \lVert x \rVert_{\Sigma_{t-1}}$, where $\alpha_{t}^{\wsb}$ is defined as in~\cref{lem::wsb::ucb}.
\end{lemma}

The \emph{noise part} accounts for the randomness in the observations and the influence of the initial prior.

\begin{lemma} \label{lem::wsb::concentration}
For any horizon $T \in \N_{+}$, $\delta \in (0,1)$, and prior $\pi(\theta) = \mathcal{N}(\theta \vert \mu_{0},\Sigma_{0})$, with probability at least $1-\delta$, the following inequalities hold for all posteriors and $t \in [T]$, simultaneously:
\begin{equation*}
    \lVert\mu_{t-1}-\bar{\mu}_{t}\rVert_{\Sigma_{t-1}^{-1}} \leq \beta_{t-1}^{\wsb}(\delta/T) + \Pi_{t-1}
\end{equation*}
and for all $x \in \cX_{t}$  we have 
\begin{align*}
\lvert \langle x, \mu_{t-1}-\bar{\mu}_{t} \rangle \rvert \leq ( \beta_{t-1}^{\wsb}(\delta/T) + \Pi_{t-1} ) \lVert x \rVert_{\Sigma_{t-1}},    
\end{align*}
where $\beta_{t}^{\wsb}(\cdot)$ and $\Pi_{t}$ are defined as in~\cref{lem::wsb::ucb}.
\end{lemma}

The proofs of \cref{lem::wsb::drift,lem::wsb::concentration} are given in \cref{sec::appendix:::wbs}, which together prove \cref{lem::wsb::ucb}. We note that in addition to isolating the dynamic penalty $\Pi_t$, our technical analysis makes a second improvement. Inspired by modern martingale techniques \citep{howard2020time}, we significantly simplify the proof of the self-normalized concentration bound (in~\cref{sec::appendix::conc_ineq_martingales}). Instead of the complex stopping-time sequences traditionally required in the linear bandit literature \citep{abbasi2011improved, russac2019weighted}, we directly leverage Ville's inequality~\citep{doob1939-ville} to establish a clean, self-normalized super-martingale structure. This allows us to elegantly isolate the noise process at any given step before taking a standard union bound over the finite horizon $T$, providing a streamlined and highly accessible alternative framework for non-stationary concentrations.

\section{Algorithms} \label{sec::algorithms}

Building on our WSB formulation in \cref{sec::wsb}, we instantiate three algorithms corresponding to widely used exploration strategies: UCB, randomized UCB, and Thompson Sampling (TS). We show that their regrets closely match or improve upon those of their WRLS-based counterparts.  Proofs are in \cref{sec::appendix:::regrets}. Pseudo-code for \texttt{WSB-LinUCB}, \texttt{WSB-RandLinUCB}, and \texttt{WSB-LinTS} in their exponential-weighted form, $w_{s,t}=\gamma^{\,t-s}$ with $\gamma\in(0,1)$, are presented in \cref{alg:wsb_linucb,alg:wsb_randlinucb,alg:wsb_lints}, respectively.

\subsection{Deterministic Exploration with WSB}

\subsubsection{Upper Confidence Bound}
Although WSB natively supports probabilistic exploration, we first instantiate a deterministic UCB algorithm to provide a direct baseline comparison against frequentist methods. 
At each round $t$, the \texttt{WSB-LinUCB} algorithm selects
\begin{equation*}
    X_{t} = \argmax_{x\in\cX_{t}} \{ \langle x, \mu_{t-1} \rangle + (\beta_{t-1}^{\wsb}(\delta/T) + \Pi_{t-1}) \lVert x \rVert_{\Sigma_{t-1}} \}.
\end{equation*}
As the Bayesian analogue to \texttt{LB-WeightUCB}, it simply replaces the WRLS estimate and fixed penalty with our posterior mean $\mu_{t-1}$ and dynamic prior term $\Pi_{t-1}$.

\begin{theorem} \label{thm::wsb::ucb::regret}
For a horizon $T\in\N_+$, any $\delta\in(0,1)$ and prior $\pi(\theta)=\mathcal{N}(\theta\vert\mu_{0},\Sigma_{0})$, with probability at least $1-2\delta$, the following inequality holds for all posteriors $\rho_{t}(\theta)=\mathcal{N}(\theta\vert\mu_{t},\Sigma_{t})$ and all $t\in[T]$ simultaneously,
\begin{align*}
R_{T} & \leq 2 L\sqrt{\lambda_{\max}(\Sigma_{0})} \sum_{t=1}^{T} \alpha_{t-1}^{\wsb} \\ & + 2^{3/2}\sigma\sqrt{CdT\Lambda_{T}} (\beta_{T}^{\wsb}(\delta/T)+\Pi_{0}),
\end{align*}
with $C=\max\{1,L^{2} \lambda_{\max}(\Sigma_{0})/\sigma^2\}$, $\Pi_{0} \leq \lVert\mu_{0}\rVert_{\Sigma_{0}^{-1}}+S\sqrt{\lambda_{\max}(\Sigma_{0}^{-1})}$ and $\Lambda_{T} = \sum_{t=1}^{T}\log(\frac{1}{w_{t-1,t}}) + \log(1 + \frac{\Tr(\Sigma_{0})L^{2}\sum_{t=1}^{T}w_{t,T}}{d\sigma^{2}})$.
\end{theorem}

\Cref{thm::wsb::ucb::regret} holds for weighting schemes $w_{s,t}$ that are non-decreasing and satisfy the multiplicative consistency condition. The following corollary exemplifies this result for the case of exponential weights.

\begin{corollary} \label{cor::wsb::ucb::regret}
Suppose $w_{s,t}=\gamma^{t-s}$. For a horizon $T\in\N_+$, any $\gamma\in(1/T,1)$ and prior $\pi(\theta)=\mathcal{N}(\theta\vert\mu_{0},\Sigma_{0})$, with probability at least $1-1/T$, the following inequality holds for all posteriors $\rho_{t}(\theta)=\mathcal{N}(\theta\vert\mu_{t},\Sigma_{t})$ and all $t\in[T]$ simultaneously,
\begin{equation*}
R_{T} \leq \tilde{\mathcal{O}}(\sqrt{d \lambda_{\max}(\Sigma_{0})} B_{T} (1-\gamma)^{-3/2} + dT\sqrt{1-\gamma}).
\end{equation*}
In particular, setting $\lambda_{\max}(\Sigma_{0})=1/d$ and $\gamma = 1 - \max\{1/T,\sqrt{B_{T}/dT}\}$ yields $R_{T} \leq \tilde\cO(d\sqrt{T})$ when the drift is mild
($B_{T} < d/T$), and $R_{T} \leq \tilde{\cO}(d^{3/4}B_{T}^{1/4}T^{3/4})$ when the drift is large ($B_T \geq d/T$).
\end{corollary}

The regret of \texttt{WSB-LinUCB} matches that of \texttt{LB-WeightUCB} while improving upon \texttt{D-LinUCB}.

\begin{algorithm}[t!]
\caption{\texttt{WSB-LinUCB} (Weighted Sequential Bayesian Upper Confidence Bound)}
\label{alg:wsb_linucb}
\begin{algorithmic}[1]
    \Require Probability $\delta\in(0,1)$, discount factor $\gamma\in(0,1)$, prior dist. $\pi(\theta)=\cN(\theta\vert\mu_{0},\Sigma_{0})$
    \State \textbf{Initialize:}  $\upsilon=\Tr(\Sigma_{0})$
    \For{each round $t\geq1$}
        \State Receive action set $\cX_{t}$
        \State Compute $\beta_{t-1}^{\wsb}(\delta/T)$ and $\Pi_{t-1}$ according to \cref{lem::wsb::ucb,lem::wsb::prior_term}, respectively
        \State \textbf{Play action} $X_{t}=\argmax_{x\in\cX_{t}} \{ \langle\mu_{t-1},x \rangle+(\beta_{t-1}^{\wsb}(\delta/T) + \Pi_{t-1}) \lVert x \rVert_{\Sigma_{t-1}} \}$ and \textbf{receive reward} $r_{t}$
        \State \textbf{Update posterior} $(\mu_t,\Sigma_t)$ according to \cref{eq:WSB-posterior-update}
    \EndFor
\end{algorithmic}
\end{algorithm}

\subsection{Randomized Exploration with WSB}

As discussed in Section 3, while frequentist algorithms must construct surrogate distributions to achieve randomized exploration, our WSB framework natively provides a full posterior covariance. The following result extends \citet[Theorem~7]{kim2020randomized} to the WSB setting, and incorporates the refined weighted analysis of \citet{wang2023revisit} to handle the drift term, yielding improved regret guarantees. The events $\cE^{\text{WSB}}$, $\cE^{\text{Conc.}}_{t}$, and $\cE^{\text{Anti-Conc.}}_{t}$ are defined in~\cref{sec::appendix:::wbs::random_exp}.

\begin{theorem} \label{thm::wsb::regret::rand_exp}
Let $p_{1},p_{2},p_{3}\in(0,1)$, a horizon $T\in\N_+$, and suppose there exists constants $c_{1},c_{2}\geq1$, which may depend on $T$ such that $\P(\cE^{\text{WSB}}) \geq 1 - p_{1}$, $\P(\cE^{\text{Conc.}}_{t}) \geq 1 - p_{2}$, and $\P(\cE^{\text{Anti-Conc.}}_{t}) \geq p_{3}$.
Then, for any prior $\pi(\theta)=\mathcal{N}(\theta\vert\mu_{0},\Sigma_{0})$, the following inequality holds for all posteriors $\rho_{t}(\theta)=\mathcal{N}(\theta\vert\mu_{t},\Sigma_{t})$ and all $t\in[T]$ simultaneously,
\begin{align*}
    R_{T} \leq & 2L\sqrt{\lambda_{\max}(\Sigma_{0})}\sum_{t=1}^{T} \alpha_{t-1}^{\wsb}
     \\ & + (c_{1}+c_{2}) \left( 1 + \frac{2}{p_{3}-p_{2}} \right) \sqrt{dT\Lambda_{T}} + T(p_{1} + p_{2}),
\end{align*}
where $\Lambda_{T}$ is as in \cref{thm::wsb::ucb::regret}.
\end{theorem}

\subsubsection{Randomized Upper Confidence Bound}

\texttt{WSB-RandLinUCB} adapts the randomized UCB principle to the Bayesian WSB setting. At each round $t$, a sample $\eta_{t}\sim\cN(0,a^{2})$ is drawn, and the action is chosen as
\begin{equation*}
X_{t} = \argmax_{x \in \cX_{t}} \{ \langle x, \mu_{t-1} \rangle + \eta_{t} \lVert x \rVert_{\Sigma_{t-1}} \}.
\end{equation*}

\begin{corollary} \label{cor::wsb::randucb::regret}
Suppose $w_{s,t}=\gamma^{t-s}$, $c_{1}=\beta_{T}^{\text{WSB}}(1/T^2)+\Pi_{0}$, $c_{2}=a\sqrt{2\log(T/2)}$, and $a^{2}=14c_{1}^{2}$. For any $\gamma\in(1/T,1)$ and prior $\pi(\theta)=\mathcal{N}(\theta\vert\mu_{0},\Sigma_{0})$, with probability at least $1-1/T$, the following inequality holds for all posteriors $\rho_{t}(\theta)=\mathcal{N}(\theta\vert\mu_{t},\Sigma_{t})$ and all $t\in[T]$ simultaneously,
\begin{equation*}
R_{T} \leq \tilde{\mathcal{O}}(\sqrt{d \lambda_{\max}(\Sigma_{0})}B_{T}(1-\gamma)^{-3/2} + dT\sqrt{1-\gamma}).
\end{equation*}
In particular, setting $\lambda_{\max}(\Sigma_{0})=1/d$ and $\gamma = 1 - \max\{1/T,\sqrt{B_{T}/dT}\}$ yields $R_{T} \leq \tilde\cO(d\sqrt{T})$ when the drift is mild
($B_{T} < d/T$), and $R_{T} \leq \tilde{\cO}(d^{3/4}B_{T}^{1/4}T^{3/4})$ when the drift is large ($B_T \geq d/T$).
\end{corollary}

\texttt{WSB-RandLinUCB} improves upon the regret of the randomized UCB algorithm \texttt{D-RandLinUCB} by a factor of $d^{1/8}$, and removes the reliance on complex \emph{local norms} by incorporating the refined analysis of \citet{wang2023revisit}. With this improvement, \texttt{WSB-RandLinUCB} achieves the same order of regret as \texttt{WSB-LinUCB} while reducing over-conservatism in the arm-selection criterion, often resulting in better empirical performance.

\begin{algorithm}[t!]
\caption{\texttt{WSB-RandLinUCB} (Weighted Sequential Bayesian Randomized Upper Confidence Bound)}
\label{alg:wsb_randlinucb}
\begin{algorithmic}[1]
    \Require Probability $\delta\in(0,1)$, discount factor $\gamma\in(0,1)$, prior dist. $\pi(\theta)=\cN(\theta\vert\mu_{0},\Sigma_{0})$, confidence level $a>0$
    \For{each round $t\geq1$}
        \State Receive action set $\cX_{t}$
        \State Randomly sample $\eta_{t}\sim\cN(0,a^{2})$
        \State \textbf{Play action} $X_{t}=\argmax_{x\in\cX_{t}}\{ \langle\mu_{t-1},x\rangle+\eta_{t}\lVert x \rVert_{\Sigma_{t-1}} \}$ and \textbf{receive reward} $r_{t}$
        \State \textbf{Update posterior} $(\mu_t,\Sigma_t)$ according to \cref{eq:WSB-posterior-update}
    \EndFor
\end{algorithmic}
\end{algorithm}

\subsubsection{Thompson Sampling}

\texttt{WSB-LinTS} implements TS by drawing a parameter vector directly from the WSB posterior and acting greedily with respect to it. 
At each round $t$, one draws $\tilde{\mu}_{t-1} = \mu_{t-1} + \Sigma_{t-1}^{1/2}\eta_{t}$ with $\eta_{t}\sim\cN(0,a^{2}\I_{d})$, and chooses
\begin{equation*}
    X_{t} = \argmax_{x \in \cX_{t}} \{ \langle x, \tilde{\mu}_{t-1} \rangle\}.
\end{equation*}

\begin{corollary} \label{cor::wsb::ts::regret}
Suppose $w_{s,t}=\gamma^{t-s}$, $c_{1}=\beta_{T}^{\text{WSB}}(1/T^2)+\Pi_{0}$, $c_{2}=a\sqrt{2\log(KT/2)}$, and $a^{2}=14c_{1}^{2}$. For any $\gamma\in(1/T,1)$ and prior $\pi(\theta)=\mathcal{N}(\theta\vert\mu_{0},\Sigma_{0})$, with probability at least $1-1/T$, the following inequality holds for all posteriors $\rho_{t}(\theta)=\mathcal{N}(\theta\vert\mu_{t},\Sigma_{t})$ and all $t\in[T]$ simultaneously,
\begin{equation*}
R_{T} \leq \tilde{\mathcal{O}}(\sqrt{d \lambda_{\max}(\Sigma_{0})}B_{T}(1-\gamma)^{-3/2} + dT\sqrt{\log(K)(1-\gamma)}).
\end{equation*}
In particular, setting $\lambda_{\max}(\Sigma_{0})=1/d$ and $\gamma = 1 - \max\{1/T,\sqrt{B_{T}/d\sqrt{\log(K)}T}\}$ yields $R_{T} \leq \tilde\cO(d\sqrt{T\log( K)})$ when the drift is mild
($B_{T} < d\sqrt{\log(K)}/T$), and $R_{T} \leq \tilde{\cO}(d^{3/4}\log(K)^{3/8}B_{T}^{1/4}T^{3/4})$ when the drift is large ($B_T \geq d\sqrt{\log(K)}/T$).
\end{corollary}

\texttt{WSB-LinTS} achieves improved regret guarantees compared to \texttt{D-LinTS}. Similarly, \texttt{WSB-LinTS} improves upon the regret of \texttt{D-LinTS} by a factor of $d^{1/8}$. Again, the theoretical improvement is mainly due to the technique by~\citet{wang2023revisit}, our WSB formulation provides a native, principled posterior for Thompson Sampling.

\begin{algorithm}[t!]
\caption{\texttt{WSB-LinTS} (Weighted Sequential Bayesian Thompson Sampling)}
\label{alg:wsb_lints}
\begin{algorithmic}[1]
    \Require Probability $\delta\in(0,1)$, discount factor $\gamma\in(0,1)$, prior dist. $\pi(\theta)=\cN(\theta\vert\mu_{0},\Sigma_{0})$, confidence level $a>0$
    \For{each round $t\geq1$}
        \State Receive action set $\cX_{t}$
        \State Randomly sample $\tilde{\mu}_{t-1}= \mu_{t-1} + \Sigma_{t-1}^{1/2}\eta_{t}$ with $\eta_{t} \sim \cN(0,a^{2}\I_{d})$
        \State \textbf{Play action} $X_{t}=\argmax_{x\in\cX_{t}}\{ \langle \tilde{\mu}_{t-1}, x \rangle \}$ and \textbf{receive reward} $r_{t}$
        \State \textbf{Update posterior} $(\mu_t,\Sigma_t)$ according to \cref{eq:WSB-posterior-update}
    \EndFor
\end{algorithmic}
\end{algorithm}

Our derived frequentist regret of $\tilde{O}(d^{3/4} B_T^{1/4} T^{3/4})$ matches the current state-of-the-art for smooth and weighted strategies in non-stationary linear bandits. While this bound is a factor of $(dT / B_T)^{1/12}$ above the theoretical lower bound of $\Omega(d^{2/3} B_T^{1/3} T^{2/3})$ established for this problem class, this gap reflects open challenge in the literature: achieving minimax optimal rates while relying solely on smooth, continuous adaptation. We note that \citet{wei2021non} successfully match this lower bound; however, their approach relies on a complex, black-box master-base algorithmic structure that requires periodic resets. In contrast, weighted strategies like WSB are highly appealing in practice exactly because they offer continuous adaptation without the need for forced restarts or fixed memory budgets. Ultimately, WSB achieves highly competitive frequentist performance for continuous methods while offering a more principled foundation for uncertainty quantification.

\section{Experiments} \label{sec::experiments}

\paragraph{Setting.}
We study two synthetic non-stationary scenarios designed to capture both abrupt and gradual changes. In both scenarios, the time horizon $T=4000$ and the reward noise $\sigma^{2}=0.15$. In both scenarios, the unknown parameter sequence $\{\theta_{t}^{*}\}$ evolves on the unit sphere in $\mathbb{R}^d$, with variation across all coordinates. Arms are sampled from the same unit sphere with $K=\lvert \cX_{t} \rvert = 48$. This ensures that \cref{ass:bounded_parameters_features} holds with $L = S = 1$. In the \emph{abruptly changing} scenario, $\{\theta_{t}^{*}\}$ is piecewise constant and switches between four spherical anchors at every $T/4$ rounds. In the \emph{slowly varying} scenario, $\{\theta_{t}^{*}\}$ moves gradually between these anchors along shortest paths on the sphere. The discount parameter $\gamma = 1 - \max\{1/T, \sqrt{B_T/dT}\}$, except for \texttt{D-LinTS} and \texttt{WSB-LinTS}, where $\gamma = 1 - \max\{1/T, \sqrt{B_T/d\sqrt{\log(K)}T}\}$. Since the sequence $\{\theta_t^{*}\}$ is known, we use the exact variation budget $B_T$.

All WSB-based algorithms use $w_{s,t}=\gamma^{t-s}$ and the same Gaussian prior $\mathcal{N}(0,\I_d)$. Following \citet{kim2020randomized}, we use a truncated Gaussian with zero mean and standard deviation $\sigma$ for both \texttt{D-RandLinUCB} and \texttt{WSB-RandLinUCB}. This ensures that the randomly sampled confidence level lies within the upper confidence bounds used by \texttt{D-LinUCB}, \texttt{LB-WeightUCB}, and \texttt{WSB-LinUCB} with high probability. As in \citet{kim2020randomized}, we adopt the non-inflated variant by setting the scaling factor $a=1$ for \texttt{D-RandLinUCB}, \texttt{WSB-RandLinUCB}, \texttt{D-LinTS}, and \texttt{WSB-LinTS}.

\begin{table*}[t!]
\caption{Cumulative regret at horizon ($T$) (mean $\pm$ std) across $100$ independent trials for varying dimensions ($d$), under both abruptly changing and slowly drifting scenarios.} 
\label{tab::experiments}
\begin{center}
\resizebox{\textwidth}{!}{%
\begin{tabular}{lccccc}
\multicolumn{6}{c}{\textbf{Abruptly Changing}} \\
\toprule
\textbf{ALGORITHM} & $d=2$ & $d=4$ & $d=6$ & $d=16$ & $d=32$ \\
\midrule
\texttt{D-LinUCB} \citep{russac2019weighted} & $455.8 \pm 21.4$ &  $713.0 \pm 73.2$ &  $867.6 \pm 90.5$ &  $1277.9 \pm 99.8$ &  $1290.0 \pm 125.9$ \\
\texttt{LB-WeightUCB} \citep{wang2023revisit} & $522.5 \pm 19.4$ &  $822.7 \pm 88.7$ &  $993.4 \pm 102.1$ &  $1360.0 \pm 106.8$ &  $1324.6 \pm 128.5$ \\
\texttt{WSB-LinUCB} (\textbf{Ours}, \cref{alg:wsb_linucb}) & $470.4 \pm 18.8$ &  $780.6 \pm 83.2$ &  $990.2 \pm 100.8$ &  $1477.9 \pm 116.0$ &  $1427.5 \pm 136.6$ \\
\hdashline
\texttt{D-RandLinUCB} \citep{kim2020randomized} & $293.8 \pm 20.6$ &  $381.7 \pm 42.9$ &  $402.4 \pm 51.7$ &  $476.4 \pm 41.8$ &  $503.6 \pm 49.9$ \\
\texttt{WSB-RandLinUCB} (\textbf{Ours}, \cref{alg:wsb_randlinucb}) & $267.5 \pm 20.7$ &  $355.0 \pm 44.9$ &  $366.5 \pm 53.9$ &  $433.7 \pm 44.1$ &  $474.2 \pm 50.7$ \\
\hdashline
\texttt{D-LinTS} \citep{kim2020randomized} & $375.8 \pm 19.6$ &  $598.8 \pm 49.2$ &  $720.3 \pm 59.9$ &  $1077.0 \pm 74.5$ &  $1138.6 \pm 101.5$ \\
\texttt{WSB-LinTS} (\textbf{Ours}, \cref{alg:wsb_lints}) & $328.9 \pm 21.2$ &  $458.9 \pm 50.4$ &  $501.3 \pm 51.5$ &  $719.5 \pm 47.8$ &  $845.5 \pm 61.0$ \\
\bottomrule
\multicolumn{6}{c}{\textbf{Slowly Drifting}} \\
\toprule
\textbf{ALGORITHM} & $d=2$ & $d=4$ & $d=6$ & $d=16$ & $d=32$ \\
\midrule
\texttt{D-LinUCB} \citep{russac2019weighted} & $440.4 \pm 23.3$ &  $674.9 \pm 167.8$ &  $857.8 \pm 167.1$ &  $1346.1 \pm 221.7$ &  $1284.0 \pm 263.5$ \\
\texttt{LB-WeightUCB} \citep{wang2023revisit} & $504.6 \pm 24.3$ &  $776.5 \pm 194.3$ &  $962.7 \pm 191.6$ &  $1405.7 \pm 234.3$ &  $1301.7 \pm 267.0$ \\
\texttt{WSB-LinUCB} (\textbf{Ours}, \cref{alg:wsb_linucb}) & $446.3 \pm 23.5$ &  $737.9 \pm 185.4$ &  $969.0 \pm 194.8$ &  $1531.2 \pm 257.7$ &  $1405.3 \pm 288.4$\\
\hdashline
\texttt{D-RandLinUCB} \citep{kim2020randomized} & $169.3 \pm 14.0$ &  $236.1 \pm 36.2$ &  $289.5 \pm 29.7$ &  $403.5 \pm 35.3$ &  $435.5 \pm 37.1$ \\
\texttt{WSB-RandLinUCB} (\textbf{Ours}, \cref{alg:wsb_randlinucb}) & $130.8 \pm 14.5$ &  $194.1 \pm 28.2$ &  $241.4 \pm 24.3$ &  $367.3 \pm 23.9$ &  $405.5 \pm 33.6$ \\
\hdashline
\texttt{D-LinTS} \citep{kim2020randomized} & $210.9 \pm 16.6$ &  $437.9 \pm 92.0$ &  $608.7 \pm 94.9$ &  $1101.1 \pm 146.8$ &  $1128.9 \pm 210.1$\\
\texttt{WSB-LinTS} (\textbf{Ours}, \cref{alg:wsb_lints}) & $136.6 \pm 13.9$ &  $268.3 \pm 49.4$ &  $368.2 \pm 45.2$ &  $702.7 \pm 63.5$ &  $840.6 \pm 107.7$ \\
\bottomrule
\end{tabular}}
\end{center}
\end{table*}

\paragraph{Results.}
The cumulative regrets across all evaluated dimensions are reported in~\cref{tab::experiments}, and the corresponding round-by-round cumulative regret trajectories are provided in~\cref{sec::regret_curves}. Across both \emph{abruptly changing} and \emph{slowly drifting} scenarios, randomized exploration substantially outperforms deterministic UCB-based exploration. Among the deterministic methods, \texttt{WSB-LinUCB} improves over \texttt{LB-WeightUCB} in lower dimensions, but becomes slightly worse as the dimension increases. However, neither \texttt{WSB-LinUCB} nor \texttt{LB-WeightUCB} outperforms \texttt{D-LinUCB}, which relies on the more complex \emph{local norm}. In contrast, the randomized WSB variants consistently improve upon their WRLS-based counterparts. \texttt{WSB-RandLinUCB} achieves lower regret than \texttt{D-RandLinUCB} across all dimensions and both scenarios, while \texttt{WSB-LinTS} substantially outperforms \texttt{D-LinTS}. These results indicate that the practical benefits of the WSB posterior are most pronounced for randomized exploration. In this case, the posterior covariance can be used directly for uncertainty-driven action selection, while the prior influence is captured through the dynamic prior term $\Pi_{t}$ rather than through a fixed worst-case initialization penalty. This yields less conservative randomized exploration without relying on surrogate posterior distributions or \emph{local norms}.

\paragraph{Ablation.} 
An ablation study is presented in~\cref{sec::ablation}, where we examine the sensitivity of the WSB-based algorithms to prior misspecification. We vary the norm of the prior mean as $\lVert\mu_0\rVert_2 \in \{1,10,100\}$, while the true parameter sequence satisfies the bound $S=1$. The results show that moderate misspecification, such as $\lVert\mu_0\rVert_2=10$, leads to a noticeable but controlled increase in regret across both abruptly changing and slowly drifting environments. In contrast, severe misspecification with $\lVert\mu_0\rVert_2=100$ can substantially degrade performance, particularly for the randomized exploration methods. This highlights that the dynamic prior term $\Pi_t$ is not merely a technical artifact: when the prior mean is far outside the true parameter scale, its influence can dominate the early posterior updates and delay adaptation. Overall, the ablation confirms that WSB can accommodate biased priors, but also emphasizes the practical importance of choosing a prior scale compatible with the assumed parameter bound $S$.

Lastly, as discussed in the introduction, GP-based methods provide a Bayesian treatment of uncertainty, but they do so through kernel representations whose size grows with the accumulated history. In long-horizon bandit problems, this growing representation can make posterior updates and action selection computationally prohibitive. WSB takes a complementary parametric route: by maintaining a Gaussian posterior over the finite-dimensional reward parameter, it retains Bayesian uncertainty quantification for both deterministic and randomized exploration while preserving efficient recursive updates.

\section{Conclusion} \label{sec::conclusion}
We introduce a new concentration inequality for WSB posteriors that incorporates prior information. Building on this result, we instantiate the WSB framework through three exploration algorithms, \texttt{WSB-LinUCB}, \texttt{WSB-RandLinUCB}, and \texttt{WSB-LinTS}, which achieve improved theoretical and empirical regrets. To support this framework, we provide a simplified proof for the time-uniform concentration of vector-valued martingales using Ville's inequality, offering a clean, alternative subroutine of independent interest for future linear bandit literature. Together, these results demonstrate that Bayesian principles, when paired with weighted updates, yield practical and theoretically sound algorithms for non-stationary sequential decision-making. While our current analysis relies on a known total variation budget $B_T$ to optimize the weighting parameter, the WSB framework natively accommodates master-base meta-tuning paradigms or online restart heuristics, making fully automated, data-driven drift adaptation a promising direction for future work.

\begin{contributions}
All authors contributed to the conceptualization and design of the research. N.~W. and Y.-S.~W. developed and refined the theoretical analysis. N.~W., .A.~A. and M.~K. designed and executed the empirical evaluations. All authors participated in the drafting and critical revision of the manuscript.
\end{contributions}

\begin{acknowledgements}
This work was supported by grants from the Novo Nordisk Foundation (NNF) under grant number NNF21OC0070621 and the Carlsberg Foundation (CF) under grant number CF21-0250. Y.-S.~W. was also supported by the Academia Sinica Postdoctoral Scholar Program, grant number AS-PD-1151-M15-2. The authors thank the anonymous reviewers for their insightful feedback, which helped improve the clarity and technical precision of the paper.
\end{acknowledgements}

\bibliography{references}

\newpage

\onecolumn

\title{Supplementary Material: Weighted Sequential Bayesian Inference for Non-Stationary Linear Contextual Bandits}
\maketitle

\appendix

\setcounter{equation}{0}
\renewcommand{\theequation}{\thesection.\arabic{equation}}
\setcounter{theorem}{0}
\renewcommand{\thetheorem}{\thesection.\arabic{theorem}}
\setcounter{lemma}{0}
\renewcommand{\thelemma}{\thesection.\arabic{lemma}}
\setcounter{proposition}{0}
\renewcommand{\theproposition}{\thesection.\arabic{proposition}}
\setcounter{corollary}{0}
\renewcommand{\thecorollary}{\thesection.\arabic{corollary}}
\setcounter{definition}{0}
\renewcommand{\thedefinition}{\thesection.\arabic{definition}}
\setcounter{assumption}{0}
\renewcommand{\theassumption}{\thesection.\arabic{assumption}}
\setcounter{example}{0}
\renewcommand{\theexample}{\thesection.\arabic{example}}
\setcounter{remark}{0}
\renewcommand{\theremark}{\thesection.\arabic{remark}}

\section{SOME TECHNICAL PROPOSITIONS} \label{sec::appendix::prop}

\begin{proposition}[Determinant inequalities] \label{prop::determinant}
For some $p\in\N_{+}$, define $\overline{\Sigma}_{t}^{-1} = \overline{\Sigma}_{0}^{-1}+\frac{1}{\sigma^{2}}\sum_{s=1}^{t}w_{s,t}^{p}X_{s}X_{s}^{\top}$, where $\overline{\Sigma}_{0}\succ0$.
Under \cref{ass:bounded_parameters_features}, we have
\begin{equation*}
\det\left(\overline{\Sigma}_{t}^{-1}\right) \leq \left(\frac{\Tr(\overline{\Sigma}_{0}^{-1}) + \frac{L^{2}}{\sigma^{2}}\sum_{s=1}^{t}w_{s,t}^{p}}{d}\right)^{d} \quad \text{and} \quad
\frac{\det\left(\overline{\Sigma}_{0}\right)}{\det\left(\overline{\Sigma}_{t}\right)} \leq \left(1 + \frac{\Tr(\overline{\Sigma}_{0})L^{2}\sum_{s=1}^{t}w_{s,t}^{p}}{d\sigma^{2}}\right)^{d}.
\end{equation*}
\end{proposition}
\begin{proof}[Proof of \cref{prop::determinant}]
Let $\lambda_{1},\dots,\lambda_{d}$ denote the eigenvalues of $\overline{\Sigma}_{t}$.
Recall that $\overline{\Sigma}_{t}$ is positive definite, thus, the eigenvalues $\lambda_{1},\dots,\lambda_{d}$ are positive.
Also, note that $\det(\overline{\Sigma}_{t}^{-1})=\prod_{i=1}^{d}\lambda_{i}^{-1}$ and $\Tr(\overline{\Sigma}_{t}^{-1})=\sum_{i=1}^{d}\lambda_{i}^{-1}$.
Hence, $\det(\overline{\Sigma}_{t}^{-1})\leq(\Tr(\overline{\Sigma}_{t}^{-1})/d)^{d}$ by the inequality of arithmetic and geometric means.
Next, since $\lVert X_{t} \rVert_{2} \leq L$ for any $t$ (\cref{ass:bounded_parameters_features}), we obtain that
\begin{align*}
\Tr(\overline{\Sigma}_{t}^{-1}) = & \Tr(\overline{\Sigma}_{0}^{-1}) + \frac{1}{\sigma^{2}}\sum_{s=1}^{t} w_{s,t}^{p}\Tr(X_{s}X_{s}^{\top}) \\
= & \Tr(\overline{\Sigma}_{0}^{-1}) + \frac{1}{\sigma^{2}}\sum_{s=1}^{t} w_{s,t}^{p}\lVert X_{s} \rVert_{2}^{2} \\ 
\leq & \Tr(\overline{\Sigma}_{0}^{-1})+\frac{L^{2}}{\sigma^{2}}\sum_{s=1}^{t}w_{s,t}^{p},
\end{align*}
which shows the first inequality of the proposition. For the second inequality, we use that
\begin{align*}
\frac{\det\left(\overline{\Sigma}_{0}\right)}{\det\left(\overline{\Sigma}_{t}\right)} = & \det\left(\overline{\Sigma}_{0}\overline{\Sigma}_{t}^{-1}\right) \\
= & \det\left(\overline{\Sigma}_{0}\left(\overline{\Sigma}_{0}^{-1}+\frac{1}{\sigma^{2}}\sum_{s=1}^{t} w_{s,t}^{p} X_{s} X_{s}^{\top} \right)\right) \\ 
= & \det\left(\mathbb{I}_{d}+\frac{1}{\sigma^{2}}\overline{\Sigma}_{0}\sum_{s=1}^{t}w_{s,t}^{p}X_{s}X_{s}^{\top}\right).
\end{align*}
Next, as $\Tr(\mathbb{I}_{d}+\frac{1}{\sigma^{2}}\overline{\Sigma}_{0}\sum_{s=1}^{t}w_{s,t}^{p}X_{s}X_{s}^{\top}) =  \Tr(\mathbb{I}_{d}) + \frac{1}{\sigma^{2}}\Tr(\overline{\Sigma}_{0}\sum_{s=1}^{t}w_{s,t}^{p}X_{s}X_{s}^{\top})$, we only need to notice that $\Tr(\overline{\Sigma}_{0}\sum_{s=1}^{t}w_{s,t}^{p}X_{s}X_{s}^{\top}) \leq \Tr(\overline{\Sigma}_{0})L^{2}\sum_{s=1}^{t}w_{s,t}^{p}$ by Cauchy-Schwarz inequality.
\end{proof}

\begin{proposition}[Mahalanobis convexity bound] \label{prop::convexity_bound}
For some $p\in\N_{+}$, define $\overline{\Sigma}_{t}^{-1} = \overline{\Sigma}_{0}^{-1}+\frac{1}{\sigma^{2}}\sum_{s=1}^{t}w_{s,t}^{p}X_{s}X_{s}^{\top}$, where $\overline{\Sigma}_{0}\succ0$. Under \cref{ass:bounded_parameters_features}, we have
\begin{equation*}
 \max_{\theta : \rVert \theta \rVert_{2} \leq S} \lVert \overline{\Sigma}_{0}^{-1}(\mu-\theta)\rVert_{\overline{\Sigma}_{t}} \leq \sqrt{ \lVert \mu \rVert_{\overline{M}_{t}}^{2} + S^2 \lambda_{\max}(\overline{M}_t) + 2 S \sqrt{\mu^\top \overline{M}_t^2 \mu} },
\end{equation*}
for any $\mu\in\R^{d}$, where $\overline{M}_{t} = \overline{\Sigma}_{0}^{-1} \overline{\Sigma}_{t} \overline{\Sigma}_{0}^{-1}$.
\end{proposition}

\begin{proof}[Proof of \cref{prop::convexity_bound}]
Let $\overline{M}_{t}=U \Lambda U^{\top}$ be an eigendecomposition with
$\Lambda=\diag(\lambda_{1},\dots,\lambda_{d})$ ordered so that $\lambda_{\max}(\overline{M}_{t})=\lambda_{1}\geq \lambda_{2}\geq\dots\geq\lambda_{d-1}\geq\lambda_{d}=\lambda_{\min}(\overline{M}_{t})>0$. Since $U$ is orthogonal ($U^\top U=UU^\top=\I_{d}$), the Euclidean norm is invariant: $\lVert x \rVert_{2} = \lVert U^\top x \rVert_{2}$. Thus, for any $\theta\in\R^{d}$,
\begin{equation*}
\lVert \mu - \theta \rVert_{\overline{M}_{t}} = \lVert \overline{M}_{t}^{1/2}(\mu-\theta) \rVert_{2} = \lVert U^\top \overline{M}_{t}^{1/2}(\mu-\theta) \rVert_{2} = \lVert U^\top U \Lambda^{1/2} U^\top(\mu-\theta) \rVert_{2} = \lVert \Lambda^{1/2} U^\top \mu - \Lambda^{1/2} U^\top \theta \rVert_{2}.
\end{equation*}
Equivalently, with $b=\overline{M}_{t}^{1/2} \mu$ (such that $U^\top b=\Lambda^{1/2}U^\top\mu$) and $z=U^\top\theta$ (such that $\lVert z \rVert_{2} = \lVert \theta \rVert_{2} \leq S$), we have
\begin{equation*}
\lVert \mu - \theta \rVert_{\overline{M}_{t}} = \lVert U^\top b - \Lambda^{1/2} z \rVert_{2}.
\end{equation*}
The map $z \mapsto \lVert U^\top b-\Lambda^{1/2}z \rVert_{2}$ is convex; hence a maximizer over the convex set $\{ z : \lVert z \rVert_{2} \leq S \}$ lies on the boundary $\lVert z \rVert_{2} = S$. Hence, we can write $z=Se$ with $\lVert e \rVert_{2} = 1$ and consider
\begin{equation*}
\phi(e) =\lVert U^{\top} b - S \Lambda^{1/2} e \rVert_{2}^{2} = \lVert b \rVert_{2}^{2} + S^{2} e^{\top} \Lambda e - 2 S b^{\top} \Lambda^{1/2} e.
\end{equation*}
Since $\Lambda \preceq \lambda_{\max}(\overline{M}_{t}) \I_{d}$ and $\lVert e \rVert_2 = 1$, we have $e^{\top}\Lambda e \leq \lambda_{\max}(\overline{M}_{t})$. For the cross term, applying the Cauchy-Schwarz inequality directly yields:
\begin{equation*}
- 2 S b^{\top} \Lambda^{1/2} e \leq 2 S \lVert \Lambda^{1/2} b \rVert_{2} \lVert e \rVert_{2} = 2 S \lVert \Lambda^{1/2} b \rVert_{2}.
\end{equation*}
Substituting these bounds back into $\phi(e)$ gives:
\begin{equation*}
\phi(e) \leq \lVert b \rVert_{2}^{2} + S^{2} \lambda_{\max}(\overline{M}_{t}) + 2 S \lVert \Lambda^{1/2} b \rVert_{2}.
\end{equation*}
Notice that $\lVert b \rVert_{2}^{2} = \mu^{\top} \overline{M}_{t} \mu = \lVert \mu \rVert_{\overline{M}_{t}}^{2}$, and for the cross-term argument, we can expand $\lVert \Lambda^{1/2} b \rVert_{2} = \sqrt{b^\top \Lambda b} = \sqrt{\mu^\top \overline{M}_t^{1/2} (U \Lambda U^\top) \overline{M}_t^{1/2} \mu} = \sqrt{\mu^\top \overline{M}_t^2 \mu}$. Taking the square root on both sides of our inequality yields:
\begin{equation*}
\lVert \mu - \theta \rVert_{\overline{M}_{t}} \leq \sqrt{ \lVert \mu \rVert_{\overline{M}_{t}}^{2} + S^2 \lambda_{\max}(\overline{M}_t) + 2 S \sqrt{\mu^\top \overline{M}_t^2 \mu} },
\end{equation*}
for every $\theta$ with $\lVert \theta \rVert_{2} \leq S$. Maximizing over the boundary completes the proof.
\end{proof}

\setcounter{equation}{0}
\renewcommand{\theequation}{\thesection.\arabic{equation}}
\setcounter{theorem}{0}
\renewcommand{\thetheorem}{\thesection.\arabic{theorem}}
\setcounter{lemma}{0}
\renewcommand{\thelemma}{\thesection.\arabic{lemma}}
\setcounter{proposition}{0}
\renewcommand{\theproposition}{\thesection.\arabic{proposition}}
\setcounter{corollary}{0}
\renewcommand{\thecorollary}{\thesection.\arabic{corollary}}
\setcounter{definition}{0}
\renewcommand{\thedefinition}{\thesection.\arabic{definition}}
\setcounter{assumption}{0}
\renewcommand{\theassumption}{\thesection.\arabic{assumption}}
\setcounter{example}{0}
\renewcommand{\theexample}{\thesection.\arabic{example}}
\setcounter{remark}{0}
\renewcommand{\theremark}{\thesection.\arabic{remark}}
\section{SELF-NORMALIZED CONCENTRATION INEQUALITY FOR VECTOR-VALUED MARTINGALES}\label{sec::appendix::conc_ineq_martingales}

In this appendix, we provide a simplified proof of the self-normalized concentration inequality for vector-valued martingales \citep{abbasi2011improved} using Ville's inequality \citep{doob1939-ville}, and provide its weighted corollary, which we rely on to bound the stochastic noise in~\cref{sec::appendix:::wbs}. 

We begin by restating the classical theorem. Let $\{\cF_t\}_{t \ge 0}$ be a filtration. Let $\{\epsilon_t\}_{t \ge 1}$ be a real-valued stochastic process such that $\epsilon_t$ is $\cF_t$-measurable and conditionally $\sigma$-sub-Gaussian, meaning that for all $\nu \in \R$, $\E[\exp(\nu \epsilon_t) \mid \cF_{t-1}] \le \exp(\nu^2 \sigma^2 / 2)$ almost surely. Let $\{X_t\}_{t \ge 1}$ be an $\R^d$-valued stochastic process where $X_t$ is $\cF_{t-1}$-measurable. Assume that $V$ is a $d \times d$ positive definite matrix. For any $t \ge 0$, define
\begin{equation*}
    \overline{V}_t = V + \sum_{s=1}^t X_s X_s^\top, \quad \text{and} \quad S_t = \sum_{s=1}^t \epsilon_s X_s.
\end{equation*}

\begin{theorem}[\citet{abbasi2011improved}, Theorem 1]\label{thm::abbasi}
For any $\delta \in (0,1)$, with probability at least $1-\delta$, the following inequality holds for all $t \ge 0$:
\begin{equation*}
    \|S_t\|_{\overline{V}_t^{-1}} \le \sigma \sqrt{2 \log\left(\frac{1}{\delta} \frac{\det(\overline{V}_t)^{1/2}}{\det(V)^{1/2}}\right)}.
\end{equation*}
\end{theorem}

For our non-stationary analysis in~\cref{sec::appendix:::wbs}, we need a bound that applies to a predictable sequence of weights $\{\kappa_t\}_{t \ge 1}$ (i.e., $\kappa_t$ is $\cF_{t-1}$-measurable), which can be obtained as a direct corollary of \cref{thm::abbasi}.

\begin{corollary}[Weighted Extension] \label{cor::weighted_concentration}
Let $\Lambda$ be a predictable positive-definite prior covariance matrix. Define $\overline{\Lambda}_t = \Lambda + \sum_{s=1}^t \kappa_s^2 X_s X_s^\top$ and $S_t = \sum_{s=1}^t \kappa_s X_s \epsilon_s$. For any $\delta \in (0,1)$, with probability at least $1-\delta$, for all $t \ge 0$,
\begin{equation*}
    \|S_t\|_{\overline{\Lambda}_t^{-1}} \le \sigma \sqrt{2 \log\left(\frac{1}{\delta} \frac{\det(\overline{\Lambda}_t)^{1/2}}{\det(\Lambda)^{1/2}}\right)}.
\end{equation*}
\end{corollary}
\begin{proof}
Because the scalar weight $\kappa_t$ is $\cF_{t-1}$-measurable, the product $\tilde{X}_t = \kappa_t X_t$ remains an $\cF_{t-1}$-measurable, predictable random vector. Directly applying \cref{thm::abbasi} by substituting $X_t$ with $\tilde{X}_t$ exactly yields the stated result.
\end{proof}

\subsection{A Simplified Proof via Ville's Inequality}

We now provide the modernized proof for the bound in \cref{thm::abbasi}. We rely on Ville's inequality, a time-uniform extension of Markov's inequality for non-negative super-martingales. A friendly proof and further discussions can be found in \citet{howard2020time}.

\begin{lemma}[Ville's Inequality] \label{lem::ville}
Let $(M_t)_{t \ge 0}$ be a non-negative super-martingale with respect to a filtration $(\cF_t)_{t \ge 0}$. For any starting time $t_0 \in \N$ and threshold $u > 0$,
\begin{equation*}
    \P(\exists t \ge t_0 : M_t \ge u) \le \frac{\E[M_{t_0}]}{u}.
\end{equation*}
\end{lemma}

\begin{proof}[Modernized Proof of \cref{thm::abbasi}]
We aim to show that with probability at least $1-\delta$, for all $t \ge 0$:
\begin{equation*}
    \|S_t\|_{\overline{V}_t^{-1}} \le \sigma \sqrt{2 \log\left(\frac{1}{\delta} \frac{\det(\overline{V}_t)^{1/2}}{\det(V)^{1/2}}\right)}.
\end{equation*}
This is equivalent to showing that with probability at most $\delta$, there exists $t \ge 0$ such that:
\begin{equation*}
    L_t \equiv \left(\frac{\det(V)}{\det(\overline{V}_t)}\right)^{1/2} \exp\left(\frac{\|S_t\|_{\overline{V}_t^{-1}}^2}{2\sigma^2}\right) \ge \frac{1}{\delta}.
\end{equation*}
If we can show that the process $(L_t)_{t \ge 0}$ is a non-negative super-martingale with initial expected value $\E[L_0] = 1$, we can apply Ville's Inequality (\cref{lem::ville}) with the threshold $u = 1/\delta$:
\begin{equation*}
    \P(\exists t \ge 0 : L_t \ge 1/\delta) \le \frac{\E[L_0]}{1/\delta} = \delta,
\end{equation*}
which would immediately imply the theorem.

We construct this super-martingale by the method of mixtures. Consider the exponential process for any fixed direction $x \in \R^d$:
\begin{equation} \label{eq::fixed_martingale}
    M_t(x) = \exp\left(\frac{1}{\sigma} x^\top S_t - \frac{1}{2} x^\top (\overline{V}_t - V) x\right).
\end{equation}
Since $X_t$ and $M_{t-1}(x)$ are $\cF_{t-1}$-measurable, we have:
\begin{align*}
    \E[M_t(x) \mid \cF_{t-1}] &= M_{t-1}(x) \exp\left(-\frac{1}{2}(x^\top X_t)^2\right) \E\left[\exp\left(\frac{\epsilon_t x^\top X_t}{\sigma}\right) \Bigm| \cF_{t-1}\right] \\
    &\le M_{t-1}(x) \exp\left(-\frac{1}{2}(x^\top X_t)^2\right) \exp\left(\frac{(x^\top X_t/\sigma)^2 \sigma^2}{2}\right) \\
    &= M_{t-1}(x),
\end{align*}
where the inequality uses the conditionally $\sigma$-sub-Gaussian property of $\epsilon_t$. Thus, $(M_t(x))_{t \ge 0}$ is a non-negative super-martingale with $M_0(x) = \exp(0) = 1$.

To control the deviation in all directions simultaneously, we integrate the fixed super-martingales over the probability density $h(x)$ of the Gaussian prior $\cN(0, V^{-1})$. The mixture process $M_t = \int_{\R^d} M_t(x) h(x) dx$ is also a non-negative super-martingale with initial value $\E[M_0] = 1$ due to Fubini's theorem:
\begin{equation*}
\E[M_t \mid \cF_{t-1}] = \int_{\R^d} \E[M_t(x) \mid \cF_{t-1}] h(x) dx \leq \int_{\R^d} M_{t-1}(x) h(x) dx = M_{t-1}.
\end{equation*}

Substituting the definitions of $M_t(x)$ and $h(x)$ into the integral:
\begin{align*}
    M_t &= \int_{\R^d} \exp\left(\frac{x^\top S_t}{\sigma} - \frac{1}{2} x^\top (\overline{V}_t - V) x\right) \frac{\det(V)^{1/2}}{(2\pi)^{d/2}} \exp\left(-\frac{1}{2} x^\top V x\right) dx \\
    &= \frac{\det(V)^{1/2}}{(2\pi)^{d/2}} \int_{\R^d} \exp\left(\frac{x^\top S_t}{\sigma} - \frac{1}{2} x^\top \overline{V}_t x\right) dx.
\end{align*}
Applying the standard Gaussian integral identity $\int \exp(b^\top x - \frac{1}{2} x^\top A x) dx = \frac{(2\pi)^{d/2}}{\det(A)^{1/2}} \exp\left(\frac{1}{2} b^\top A^{-1} b\right)$ with $A = \overline{V}_t$ and $b = S_t/\sigma$, we obtain:
\begin{equation*}
    M_t = \frac{\det(V)^{1/2}}{(2\pi)^{d/2}} \cdot \frac{(2\pi)^{d/2}}{\det(\overline{V}_t)^{1/2}} \exp\left( \frac{1}{2} \left(\frac{S_t}{\sigma}\right)^\top \overline{V}_t^{-1} \left(\frac{S_t}{\sigma}\right) \right) = \left(\frac{\det(V)}{\det(\overline{V}_t)}\right)^{1/2} \exp\left(\frac{\|S_t\|_{\overline{V}_t^{-1}}^2}{2\sigma^2}\right).
\end{equation*}
Therefore, $M_t = L_t$. Applying Ville's inequality to this mixture process completes the proof.
\end{proof}

\setcounter{equation}{0}
\renewcommand{\theequation}{\thesection.\arabic{equation}}
\setcounter{theorem}{0}
\renewcommand{\thetheorem}{\thesection.\arabic{theorem}}
\setcounter{lemma}{0}
\renewcommand{\thelemma}{\thesection.\arabic{lemma}}
\setcounter{proposition}{0}
\renewcommand{\theproposition}{\thesection.\arabic{proposition}}
\setcounter{corollary}{0}
\renewcommand{\thecorollary}{\thesection.\arabic{corollary}}
\setcounter{definition}{0}
\renewcommand{\thedefinition}{\thesection.\arabic{definition}}
\setcounter{assumption}{0}
\renewcommand{\theassumption}{\thesection.\arabic{assumption}}
\setcounter{example}{0}
\renewcommand{\theexample}{\thesection.\arabic{example}}
\setcounter{remark}{0}
\renewcommand{\theremark}{\thesection.\arabic{remark}}
\section{A BAYESIAN TREATMENT OF NON-STATIONARY LINEAR CONTEXTUAL BANDITS} \label{sec::appendix:::wbs}

This appendix provides the detailed proofs for the results stated in \cref{sec::wsb}, which establish high-probability confidence bounds for the WSB posteriors. In particular, \cref{lem::wsb::ucb} establishes a uniform deviation inequality for Bayesian posteriors over a finite horizon $T$. Unlike prior work, our analysis avoids the introduction of an auxiliary covariance matrix by building on the refined framework of \citet{wang2023revisit}.

We structure the appendix as follows: We begin by analyzing the two components of the estimation error separately --- namely, the effect of parameter drift (\cref{lem::wsb::drift}) and stochastic noise including prior influence (\cref{lem::wsb::concentration}). We then combine these results via a finite-horizon union bound to obtain the full posterior confidence bound stated in \cref{lem::wsb::ucb}. Finally, we prove the upper bounds on the prior term (\cref{lem::wsb::prior_term}).

\begin{proof}[Proof of \cref{lem::wsb::drift}]
We aim to bound the drift-induced deviation between the surrogate posterior mean $\bar{\mu}_t$ in \cref{eq::posterior::surrogate::mean} and the current reward parameter $\theta_{t}^{*}$:
\begin{equation*}
\bar{\mu}_{t}-\theta_{t}^{*} = \Sigma_{t-1}\left(\Sigma_{0}^{-1}\theta_{t}^{*}+\frac{1}{\sigma^{2}}\sum_{s=1}^{t-1}w_{s,t-1}X_{s}X_{s}^{\top}\theta_{s}^{*} \right) - \theta_{t}^{*}.
\end{equation*}
Using the definition of the posterior covariance $\Sigma_{t-1}^{-1}$ from \cref{eq::posterior::cov} gives
\begin{align*}
\bar{\mu}_{t}-\theta_{t}^{*} &= \Sigma_{t-1}\left(\Sigma_{0}^{-1}\theta_{t}^{*}+\frac{1}{\sigma^{2}}\sum_{s=1}^{t-1}w_{s,t-1}X_{s}X_{s}^{\top}\theta_{s}^{*} - \Sigma_{t-1}^{-1} \theta_{t}^{*}\right)
\\ &= \Sigma_{t-1}\left(\Sigma_{0}^{-1}\theta_{t}^{*}+\frac{1}{\sigma^{2}}\sum_{s=1}^{t-1}w_{s,t-1}X_{s}X_{s}^{\top}\theta_{s}^{*} - \Sigma_{0}^{-1}\theta_{t}^{*}- \frac{1}{\sigma^{2}}\sum_{s=1}^{t-1}w_{s,t-1}X_{s}X_{s}^{\top}\theta_{t}^{*}\right)
\\ &= \Sigma_{t-1}\left(\frac{1}{\sigma^{2}}\sum_{s=1}^{t-1}w_{s,t-1}X_{s}X_{s}^{\top}(\theta_{s}^{*} -\theta_{t}^{*})\right).
\end{align*}
Now, for any arbitrary vector $x\in\R^{d}$, we have
\begin{equation*}
\left\langle x, \bar{\mu}_{t}-\theta_{t}^{*} \right\rangle = \left\langle x , \Sigma_{t-1} \left( \frac{1}{\sigma^{2}}\sum_{s=1}^{t-1}w_{s,t-1}X_{s}X_{s}^{\top}(\theta_{s}^{*}-\theta_{t}^{*}) \right) \right\rangle,
\end{equation*}
which by Cauchy-Schwarz inequality yields the following inequality:
\begin{equation*}
\vert\langle x, \bar{\mu}_{t}-\theta_{t}^{*}\rangle\vert \leq  \lVert x \rVert_{\Sigma_{t-1}} \left\lVert\frac{1}{\sigma^{2}}\sum_{s=1}^{t-1}w_{s,t-1}X_{s}X_{s}^{\top}(\theta_{s}^{*}-\theta_{t}^{*}) \right\rVert_{\Sigma_{t-1}}.
\end{equation*}
We now focus on bounding the right-hand norm. We first expand the inner sum over variations (i.e., $\theta_{s}^{*}-\theta_{t}^{*}=\sum_{k=s}^{t-1}(\theta_{k}^{*}-\theta_{k+1}^{*})$), swap the order of summation, apply the triangle inequality, use the Cauchy-Schwarz inequality, invoke \cref{ass:bounded_parameters_features}, and then apply the triangle inequality once more:
\begin{align*}
\left\lVert\frac{1}{\sigma^{2}}\sum_{s=1}^{t-1}w_{s,t-1}X_{s}X_{s}^{\top}(\theta_{s}^{*}-\theta_{t}^{*}) \right\rVert_{\Sigma_{t-1}}
 & = \left\lVert\frac{1}{\sigma^{2}}\sum_{s=1}^{t-1}w_{s,t-1}X_{s}X_{s}^{\top}\left(\sum_{k=s}^{t-1}(\theta_{k}^{*}-\theta_{k+1}^{*}) \right)\right\rVert_{\Sigma_{t-1}}
\\ & = \left\lVert\sum_{k=1}^{t-1}\left(\frac{1}{\sigma^{2}}\sum_{s=1}^{k}w_{s,t-1}X_{s}X_{s}^{\top}(\theta_{k}^{*}-\theta_{k+1}^{*})\right)\right\rVert_{\Sigma_{t-1}}
\\ & \leq \sum_{k=1}^{t-1} \left\lVert \frac{1}{\sigma^{2}}\sum_{s=1}^{k}w_{s,t-1}X_{s}X_{s}^{\top} ( \theta_{k}^{*}-\theta_{k+1}^{*} ) \right\rVert_{\Sigma_{t-1}}
\\ & \leq \sum_{k=1}^{t-1} \left\lVert \frac{1}{\sigma^{2}}\sum_{s=1}^{k}w_{s,t-1}X_{s}\lVert X_{s} \rVert_{2} \lVert \theta_{k}^{*}-\theta_{k+1}^{*} \rVert_{2} \right\rVert_{\Sigma_{t-1}} 
\\ & \leq L \sum_{k=1}^{t-1} \left\lVert \frac{1}{\sigma^{2}}\sum_{s=1}^{k}w_{s,t-1}X_{s} \lVert \theta_{k}^{*}-\theta_{k+1}^{*} \rVert_{2} \right\rVert_{\Sigma_{t-1}} 
\\ & \leq L \sum_{k=1}^{t-1} \frac{1}{\sigma^{2}} \sum_{s=1}^{k} w_{s,t-1}\left\lVert X_{s} \right\rVert_{\Sigma_{t-1}}  \lVert \theta_{k}^{*}-\theta_{k+1}^{*} \rVert_{2}.
\end{align*}
Applying Cauchy-Schwarz to the inner sum yields
\begin{align*}
\frac{1}{\sigma^{2}} \sum_{s=1}^{k} w_{s,t-1}\left\lVert X_{s} \right\rVert_{\Sigma_{t-1}} & = \frac{1}{\sigma^{2}} \sum_{s=1}^{k} \sqrt{w_{s,t-1}} \sqrt{w_{s,t-1}} \left\lVert X_{s} \right\rVert_{\Sigma_{t-1}} \\ & \leq \sqrt{\sum_{s=1}^{k} w_{s,t-1}} \sqrt{\frac{1}{\sigma^{2}} \sum_{s=1}^{k} w_{s,t-1} \left\lVert X_{s} \right\rVert_{\Sigma_{t-1}}^{2}} \\ & \leq \sqrt{d} \sqrt{\sum_{s=1}^{k} w_{s,t-1}},
\end{align*}
where we bounded the second factor as follows:
\begin{align*}
\frac{1}{\sigma^{2}}\sum_{s=1}^{k} w_{s,t-1} \lVert X_{s} \rVert_{\Sigma_{t-1}}^{2} & = \frac{1}{\sigma^{2}}\sum_{s=1}^{k} w_{s,t-1} \Tr\left(X_{s}^{\top}\Sigma_{t-1}X_{s}\right) 
\\ & = \Tr\left( \Sigma_{t-1} \left(\frac{1}{\sigma^{2}}\sum_{s=1}^{k} w_{s,t-1}X_{s}X_{s}^{\top}\right)\right)
\\ & \leq \Tr\left( \Sigma_{t-1} \left(\frac{1}{\sigma^{2}}\sum_{s=1}^{k} w_{s,t-1}X_{s}X_{s}^{\top}\right)\right) + \Tr\left( \Sigma_{t-1} \left(\frac{1}{\sigma^{2}}\sum_{s=k+1}^{t-1} w_{s,t-1}X_{s}X_{s}^{\top}\right)\right) + \Tr\left( \Sigma_{t-1}\Sigma_{0}^{-1}\right)
\\ & = \Tr\left( \Sigma_{t-1} \left(\Sigma_{0}^{-1} + \frac{1}{\sigma^{2}}\sum_{s=1}^{t-1} w_{s,t-1}X_{s}X_{s}^{\top}\right)\right)
\\ & = \Tr\left( \Sigma_{t-1} \Sigma_{t-1}^{-1}\right) = \Tr(\mathbb{I}_{d}) = d.
\end{align*}
By putting everything together, we obtain:
\begin{equation*}
\left\lVert\frac{1}{\sigma^{2}}\sum_{s=1}^{t-1}w_{s,t-1}X_{s}X_{s}^{\top}(\theta_{s}^{*}-\theta_{t}^{*}) \right\rVert_{\Sigma_{t-1}} \leq \sqrt{d} L \sum_{k=1}^{t-1} \sqrt{\sum_{s=1}^{k} w_{s,t-1}} \lVert \theta_{k}^{*}-\theta_{k+1}^{*} \rVert_{2}.
\end{equation*}
Combining all bounds proves that $\vert\langle x, \bar{\mu}_{t}-\theta_{t}^{*}\rangle\vert \leq \alpha_{t-1}^{\wsb} \lVert x \rVert_{\Sigma_{t-1}}$, and taking $x=\Sigma_{t-1}^{-1}(\bar{\mu}_{t}-\theta_{t}^{*})$ yields $\lVert \bar{\mu}_{t}-\theta_{t}^{*}\rVert_{\Sigma_{t-1}^{-1}} \leq \alpha_{t-1}^{\wsb}$.
\end{proof}

\begin{proof}[Proof of \cref{lem::wsb::concentration}]
We aim to bound the deviation between the posterior mean $\mu_{t-1}$ in \cref{eq::posterior::mean} and the surrogate posterior mean $\bar{\mu}_t$ in \cref{eq::posterior::surrogate::mean} with high probability:
\begin{align*}
\mu_{t-1} - \bar{\mu}_{t} & = \Sigma_{t-1}\left(\Sigma_{0}^{-1}\mu_{0}+\frac{1}{\sigma^{2}}\sum_{s=1}^{t-1}w_{s,t-1}X_{s}r_{s} \right) - \Sigma_{t-1}\left(\Sigma_{0}^{-1}\theta_{t}^{*}+\frac{1}{\sigma^{2}}\sum_{s=1}^{t-1}w_{s,t-1}X_{s}X_{s}^{\top}\theta_{s}^{*} \right) 
\\ & = \Sigma_{t-1}\left(\Sigma_{0}^{-1}(\mu_{0}-\theta_{t}^{*})+\frac{1}{\sigma^{2}}\sum_{s=1}^{t-1}w_{s,t-1}X_{s}\varepsilon_{s} \right),
\end{align*}
by the definition of the rewards $r_{t}$. For any arbitrary vector $x\in\R^{d}$, we can express the equality above as
\begin{align*}
\left\langle x,\mu_{t-1}-\bar{\mu}_{t}\right\rangle & = \left\langle x,\Sigma_{t-1}\Sigma_{0}^{-1}(\mu_{0}-\theta_{t}^{*})\right\rangle + \left\langle x,\Sigma_{t-1} \left( \frac{1}{\sigma^{2}} \sum_{s=1}^{t-1} w_{s,t-1} X_{s} \varepsilon_{s} \right)\right\rangle 
\\ & = \left\langle x,\Sigma_{0}^{-1}(\mu_{0}-\theta_{t}^{*})\right\rangle_{\Sigma_{t-1}} + \left\langle x, \frac{1}{\sigma^{2}} \sum_{s=1}^{t-1} w_{s,t-1} X_{s} \varepsilon_{s} \right\rangle_{\Sigma_{t-1}},
\end{align*}
which, by the Cauchy-Schwarz inequality, yields
\begin{equation*}
\lvert \langle x,\mu_{t-1}-\bar{\mu}_{t}\rangle \vert \leq \lVert x \rVert_{\Sigma_{t-1}} \left( \left\lVert \Sigma_{0}^{-1}(\mu_{0}-\theta_{t}^{*})\right\rVert_{\Sigma_{t-1}} + \left\lVert \frac{1}{\sigma^{2}} \sum_{s=1}^{t-1} w_{s,t-1} X_{s} \varepsilon_{s} \right\rVert_{\Sigma_{t-1}} \right).
\end{equation*}
Now we can bound each term inside the parenthesis. For the first term, we simply use \cref{ass:bounded_parameters_features} to obtain our prior term:
\begin{equation*}
 \left\lVert \Sigma_{0}^{-1}(\mu_{0}-\theta_{t}^{*})\right\rVert_{\Sigma_{t-1}} \leq \max_{\theta:\lVert\theta\rVert_{2}\leq S}\lVert \Sigma_{0}^{-1}(\mu_{0}-\theta)\rVert_{\Sigma_{t-1}} \equiv \Pi_{t-1},
\end{equation*}
which we further bound in \cref{lem::wsb::prior_term}. Next, the second term is a martingale term, which can be bounded with high probability using a weighted variant of self-normalized concentration inequality for vector-valued martingales (\cref{sec::appendix::conc_ineq_martingales}). For this analysis, we introduce the auxiliary covariance matrix $\widetilde{\Sigma}_{t}$, defined as $\widetilde{\Sigma}_{t}^{-1} = \Sigma_{0}^{-1} + \frac{1}{\sigma^{2}}\sum_{s=1}^{t} w_{s,t}^{2}X_{s}X_{s}^{\top}$, which is also positive semi-definite.
Since $\{w_{s,t} \in [0,1] : 1 \leq s \leq t\}$, and hence, $w_{s,t}\ge w_{s,t}^2$, we have $\Sigma_{t}^{-1} = \Sigma_{0}^{-1} + \frac{1}{\sigma^{2}}\sum_{s=1}^{t} w_{s,t}X_{s}X_{s}^{\top} \succeq \Sigma_{0}^{-1} + \frac{1}{\sigma^{2}}\sum_{s=1}^{t} w_{s,t}^{2}X_{s}X_{s}^{\top} = \widetilde{\Sigma}_{t}^{-1}$.
Importantly, $\widetilde{\Sigma}_{t}$ is purely used for analysis and is not required in our algorithm. 
Because $A\succeq B \Rightarrow A^{-1}\preceq B^{-1}$ for positive definite matrices, we get $\Sigma_t \preceq \widetilde{\Sigma}_t$. Hence, %
\begin{equation*}
\left\lVert \sum_{s=1}^{t-1} w_{s,t-1} X_{s} \varepsilon_{s} \right\rVert_{\Sigma_{t-1}} \leq \left\lVert \sum_{s=1}^{t-1} w_{s,t-1} X_{s} \varepsilon_{s} \right\rVert_{\widetilde{\Sigma}_{t-1}}.
\end{equation*}
Thus, we can apply \cref{cor::weighted_concentration} alongside a union bound over $t \in [T]$ with a localized failure probability $\delta_t = \delta/T$ for each time step. This guarantees that with probability at least $1-\delta$, the following inequality holds simultaneously for all $t \in [T]$:
\begin{equation*}
\left\lVert \sum_{s=1}^{t-1} w_{s,t-1} X_{s} \varepsilon_{s} \right\rVert_{\widetilde{\Sigma}_{t-1}} \leq \sigma\sqrt{2\log\left(\frac{T}{\delta}\sqrt{\frac{\det(\widetilde{\Sigma}_{t-1}^{-1})}{\det(\Sigma_{0}^{-1})}}\right)} = \sigma\sqrt{2\log\left(\frac{T}{\delta}\sqrt{\frac{\det(\Sigma_{0})}{\det(\widetilde{\Sigma}_{t-1})}}\right)}.
\end{equation*}
Applying \cref{prop::determinant} with $p=2$, gives
\begin{equation*}
\frac{\det(\Sigma_{0})}{\det(\widetilde{\Sigma}_{t-1})} \leq \left(1 + \frac{\Tr(\Sigma_{0})L^{2}\sum_{s=1}^{t-1}w_{s,t-1}^{2}}{d\sigma^{2}}\right)^{d}.
\end{equation*}
Summing everything together gives that with probability at least $1-\delta$, for all $t\in[T]$ simultaneously:
\begin{equation*}
\lvert \langle x,\mu_{t-1}-\bar{\mu}_{t}\rangle \vert \leq \lVert x \rVert_{\Sigma_{t-1}} \left( \Pi_{t-1} + \beta_{t-1}^{\wsb}(\delta/T) \right).
\end{equation*}
At last, taking $x=\Sigma_{t-1}^{-1}(\mu_{t-1}-\bar{\mu}_{t})$ and applying the dual norm inequality yields that with probability at least $1-\delta$, for all $t\in[T]$ simultaneously:
\begin{equation*}
\lVert\mu_{t-1}-\bar{\mu}_{t}\rVert_{\Sigma_{t-1}^{-1}} \leq \Pi_{t-1} + \beta_{t-1}^{\wsb}(\delta/T).
\end{equation*}
\end{proof}

\begin{proof}[Proof of \cref{lem::wsb::ucb}]
The result follows directly by combining the decomposition
\begin{equation*}
\mu_{t-1} - \theta_t^* = (\mu_{t-1} - \bar{\mu}_t) + (\bar{\mu}_t - \theta_t^*),
\end{equation*}
with the uniform bounds from \cref{lem::wsb::concentration} and \cref{lem::wsb::drift}, respectively. Specifically, for any $x \in \R^d$, the projection bound follows explicitly by applying the dual norm inequality $|\langle x,z\rangle| \leq \|x\|_{\Sigma_{t-1}}\|z\|_{\Sigma_{t-1}^{-1}}$ to the self-normalized vector error. Thus, with probability at least $1 - \delta$, we have for all $t \in [T]$ simultaneously:
\begin{equation*}
\lvert \langle x, \mu_{t-1} - \theta_t^* \rangle \rvert \leq \lvert \langle x, \mu_{t-1} - \bar{\mu}_t \rangle \rvert + \lvert\langle x, \bar{\mu}_t - \theta_t^* \rangle \rvert \leq \lVert x\rVert_{\Sigma_{t-1}} \left( \Pi_{t-1} + \beta_{t-1}^{\wsb}(\delta/T) + \alpha_{t-1}^{\wsb} \right),
\end{equation*}
where $\Pi_{t-1}$ is the prior term, $\beta_{t-1}^{\wsb}(\delta/T)$ is the noise concentration term, and $\alpha_{t-1}^{\wsb}$ is the drift bound. Finally, taking $x = \Sigma_{t-1}^{-1}(\mu_{t-1} - \theta_t^*)$ yields
\begin{equation*}
\|\mu_{t-1} - \theta_t^*\|_{\Sigma_{t-1}^{-1}} \leq \Pi_{t-1} + \beta_{t-1}^{\wsb}(\delta/T) + \alpha_{t-1}^{\wsb},
\end{equation*}
with probability at least $1 - \delta$, concluding the proof.
\end{proof}

\begin{proof}[Proof of \cref{lem::wsb::prior_term}]
The bound $\Pi_{t}^{\cxv}$ follows directly from applying \cref{prop::convexity_bound} under $p=1$ and replacing variables with the corresponding prior metrics:
\begin{equation*}
\Pi_{t} \leq \sqrt{\lVert \mu_{0} \rVert_{M_{t}}^{2} + S^2 \lambda_{\max}(M_t) + 2 S \sqrt{\mu_0^\top M_t^2 \mu_0}} \equiv \Pi_{t}^{\cxv}.
\end{equation*}
The $\Pi_{t}^{\Delta}$-bound comes from applying the triangle inequality:
\begin{equation*}
\Pi_{t} \leq \lVert \Sigma_{0}^{-1}\mu_{0} \rVert_{\Sigma_{t}} + S \lVert \Sigma_{0}^{-1} \rVert_{\Sigma_{t}} \leq  \lVert \Sigma_{0}^{-1}\mu_{0} \rVert_{\Sigma_{t}} + S \sqrt{\lambda_{\max}( \Sigma_{0}^{-1}\Sigma_{t}\Sigma_{0}^{-1})} = \lVert \mu_{0} \rVert_{M_{t}} + S \sqrt{\lambda_{\max}(M_{t})} \equiv \Pi_{t}^{\Delta},
\end{equation*}
with $M_{t} = \Sigma_0^{-1}\Sigma_{t}\Sigma_0^{-1}$.
Finally, to see that $\Pi_t^{\cxv} \leq \Pi_t^{\Delta}$, note that by the definition of the spectral norm, $\sqrt{\mu_0^\top M_t^2 \mu_0} \leq \sqrt{\lambda_{\max}(M_t) \mu_0^\top M_t \mu_0} = \sqrt{\lambda_{\max}(M_t)} \|\mu_0\|_{M_t}$. Substituting this into $(\Pi_t^{\cxv})^2$ reveals a perfect square expanding inequality:
\begin{equation*}
(\Pi_t^{\cxv})^2 \leq \|\mu_0\|_{M_t}^2 + S^2 \lambda_{\max}(M_t) + 2 S \sqrt{\lambda_{\max}(M_t)} \|\mu_0\|_{M_t} = \left(\|\mu_0\|_{M_t} + S \sqrt{\lambda_{\max}(M_t)}\right)^2 = (\Pi_t^{\Delta})^2.
\end{equation*}
Taking the square root on both sides confirms $\Pi_t^{\cxv} \leq \Pi_t^{\Delta}$, completing the proof.
\end{proof}

\setcounter{equation}{0}
\renewcommand{\theequation}{\thesection.\arabic{equation}}
\setcounter{theorem}{0}
\renewcommand{\thetheorem}{\thesection.\arabic{theorem}}
\setcounter{lemma}{0}
\renewcommand{\thelemma}{\thesection.\arabic{lemma}}
\setcounter{proposition}{0}
\renewcommand{\theproposition}{\thesection.\arabic{proposition}}
\setcounter{corollary}{0}
\renewcommand{\thecorollary}{\thesection.\arabic{corollary}}
\setcounter{definition}{0}
\renewcommand{\thedefinition}{\thesection.\arabic{definition}}
\setcounter{assumption}{0}
\renewcommand{\theassumption}{\thesection.\arabic{assumption}}
\setcounter{example}{0}
\renewcommand{\theexample}{\thesection.\arabic{example}}
\setcounter{remark}{0}
\renewcommand{\theremark}{\thesection.\arabic{remark}}
\section{REGRET GUARANTEES OF ALGORITHMS} \label{sec::appendix:::regrets}

This appendix establishes regret guarantees for our WSB algorithms, treating each exploration paradigm separately. In the case of \emph{deterministic exploration} (\cref{sec::appendix:::wbs::det_exp}), we prove the regret bound for \texttt{WSB-LinUCB} (\cref{sec::appendix:::wbs::ucb}). 
For \emph{randomized exploration} (\cref{sec::appendix:::wbs::random_exp}), we first develop the auxiliary events and instantaneous-regret bounds needed for perturbation-based policies, before deriving algorithm-specific guarantees for \texttt{WSB-RandLinUCB} (\cref{sec::appendix:::wbs::randucb}) and \texttt{WSB-LinTS} (\cref{sec::appendix:::wbs::ts}).

We begin with a technical lemma that controls the cumulative variance terms, $\sum_{t=1}^{T} \lVert X_{t} \rVert_{\Sigma_{t-1}}^{2}$. In WRLS-based approaches, this term is typically achieved using auxiliary results, such as the \emph{weighted potential lemmas} presented in \citet[Lemma~8]{faury2021regret} and/or \citet[Lemma~11]{wang2023revisit}. Here, we extend this type of control to the WSB posterior for any weighted scheme $\{w_{s,t}\}$ that are non-decreasing in $s$ (i.e., $w_{s-1,t} \leq w_{s,t}$) and satisfy the multiplicative consistency condition (i.e., $w_{s,t} \geq w_{t-1,t}  w_{s,t-1}$) in the unit interval.

\begin{lemma}[Weighted potential lemma] \label{lem::norm_bound}
For some $p\in\N_{+}$, define $\overline{\Sigma}_{t}^{-1} = \overline{\Sigma}_{0}^{-1}+\frac{1}{\sigma^{2}}\sum_{s=1}^{t}w_{s,t}^{p}X_{s}X_{s}^{\top}$, where $\overline{\Sigma}_{0}\succ0$.
Under \cref{ass:bounded_parameters_features}, we have
\begin{equation*}
\sum_{t=1}^{T}\lVert X_{t} \rVert^{2}_{\overline{\Sigma}_{t-1}} \leq 2\sigma^{2}\max\{1,L^{2} \lambda_{\max}(\overline{\Sigma}_{0})/\sigma^2\}\left( d\sum_{t=1}^{T}\log\left(\frac{1}{w_{t-1,t}^{p}}\right) + \log\left(\frac{\det(\overline{\Sigma}_{0})}{\det(\overline{\Sigma}_{T})}\right) \right).
\end{equation*}
\end{lemma}

\begin{proof}[Proof of \cref{lem::norm_bound}]

First, since $\{w_{s,t}\}$ is non-decreasing and satisfies the multiplicative consistency condition, we have $w_{s,t}^{p}\ge w_{t-1,t}^{p}\,w_{s,t-1}^{p}$ for all $s\le t-1$ (equality for exponential
weights) and $w_{t,t}^{p}\ge w_{t-1,t}^{p}$. Hence,
\begin{align*}
    \overline{\Sigma}_{t}^{-1}
    &= \overline{\Sigma}_{0}^{-1}
      +\frac{1}{\sigma^{2}}\!\left(\sum_{s=1}^{t-1} w_{s,t}^{p} X_{s}X_{s}^{\top}
      + w_{t,t}^{p} X_{t}X_{t}^{\top}\right)\\
    &\succeq w_{t-1,t}^{p}\!\left(
        \overline{\Sigma}_{0}^{-1}
        +\frac{1}{\sigma^{2}}\sum_{s=1}^{t-1} w_{s,t-1}^{p} X_{s}X_{s}^{\top}
        +\frac{1}{\sigma^{2}} X_{t}X_{t}^{\top}\right)\\
    &= w_{t-1,t}^{p}\!\left(\overline{\Sigma}_{t-1}^{-1}
        +\frac{1}{\sigma^{2}} X_{t}X_{t}^{\top}\right).
\end{align*}
Moreover, for any positive semi-definite matrices $A,B$,
$A+B=A^{1/2}\!\left(I+A^{-1/2} B A^{-1/2}\right)\!A^{1/2}$. Applying this with
$A=\overline{\Sigma}_{t-1}^{-1}$ and $B=\frac{1}{\sigma^{2}}X_t X_t^{\top}$ yields
\[
\overline{\Sigma}_{t-1}^{-1}+\frac{1}{\sigma^{2}}X_{t}X_{t}^{\top}
=\overline{\Sigma}_{t-1}^{-1/2}\!\left(
\mathbb{I}_{d}+\frac{1}{\sigma^{2}}\overline{\Sigma}_{t-1}^{1/2} X_{t}X_{t}^{\top}\overline{\Sigma}_{t-1}^{1/2}
\right)\!\overline{\Sigma}_{t-1}^{-1/2}.
\]

Next, taking the determinant on both sides gives us
\begin{align*}
\det\left(\overline{\Sigma}_{t}^{-1}\right) & \geq \det\left( w_{t-1,t}^{p} \overline{\Sigma}_{t-1}^{-1} \right) \det\left( \mathbb{I}_{d}+\frac{1}{\sigma^{2}}\left( \overline{\Sigma}_{t-1}^{1/2}X_{t} \right) \left( \overline{\Sigma}_{t-1}^{1/2}X_{t} \right)^{\top} \right)
\\ & = (w_{t-1,t}^{p})^{d} \det\left( \overline{\Sigma}_{t-1}^{-1} \right) \det\left( \mathbb{I}_{d}+\frac{1}{\sigma^{2}}\left( \overline{\Sigma}_{t-1}^{1/2}X_{t} \right) \left( \overline{\Sigma}_{t-1}^{1/2}X_{t} \right)^{\top} \right)
\\ & = (w_{t-1,t}^{p})^{d} \det\left( \overline{\Sigma}_{t-1}^{-1} \right) \left(1+\frac{1}{\sigma^{2}} \lVert X_{t} \rVert^{2}_{\overline{\Sigma}_{t-1}} \right)
\\ & \geq (w_{t-1,t}^{p})^{d} \det\left( \overline{\Sigma}_{t-1}^{-1} \right) \left(1+\frac{1}{\sigma^{2}\max\{1,L^{2} \lambda_{\max}(\overline{\Sigma}_{0})/\sigma^2\}} \lVert X_{t} \rVert^{2}_{\overline{\Sigma}_{t-1}} \right)
\\ & \geq (w_{t-1,t}^{p})^{d} \det\left( \overline{\Sigma}_{t-1}^{-1} \right) \exp\left( \frac{1}{2\sigma^{2}\max\{1,L^{2} \lambda_{\max}(\overline{\Sigma}_{0})/\sigma^2\}}\lVert X_{t} \rVert^{2}_{\overline{\Sigma}_{t-1}} \right),
\end{align*}
using $\det(1+xx^{\top})=1+\lVert x \rVert_{2}^{2}$ for any $x\in\R^{d}$, $\max\{1,L^{2} \lambda_{\max}(\overline{\Sigma}_{0})/\sigma^2\}\ge 1$, and $1+z \geq \exp(z/2)$ for any $z\in[0,1]$. Note that the term $C=\max\{1,L^{2} \lambda_{\max}(\overline{\Sigma}_{0})/\sigma^2\}$ is used in the denominator so that $\frac{1}{\sigma^{2}C}\|X_t\|_{\overline{\Sigma}_{t-1}}^{2}\in[0,1]$. This is due to the fact that $\overline{\Sigma}_{t-1}\preceq\overline{\Sigma}_{0}$ and $\|X_t\|_2\le L$ together imply
$\|X_t\|_{\overline{\Sigma}_{t-1}}^{2}
\le \lambda_{\max}(\overline{\Sigma}_{t-1})\|X_t\|_2^2
\le L^{2}\lambda_{\max}(\overline{\Sigma}_{0})$.
Finally, by exploiting the telescoping structure of the inequality above, we can derive the desired result.
\end{proof}

\subsection{Deterministic Exploration with WSB Confidence Bounds} \label{sec::appendix:::wbs::det_exp}

\subsubsection{Upper Confidence Bound (\texttt{WSB-LinUCB})} \label{sec::appendix:::wbs::ucb}

\begin{proof}[Proof of \cref{thm::wsb::ucb::regret}]
For any $x\in\R^{d}$, by positive definiteness, we have $\lVert x \rVert_{\Sigma_{t-1}} = \sqrt{x^\top \Sigma_{t-1} x} \leq \sqrt{\lambda_{\max}(\Sigma_{t-1})} \lVert x \rVert_{2}$. From the update $\Sigma_{t-1}^{-1}=\Sigma_0^{-1}+\frac{1}{\sigma^2}\sum_{s=1}^{t-1} w_{s,t-1} X_sX_s^\top \succeq \Sigma_0^{-1}$, we have $\Sigma_{t-1}\preceq \Sigma_0$, hence $\lambda_{\max}(\Sigma_{t-1}) \leq \lambda_{\max}(\Sigma_0)$. Therefore, using \cref{ass:bounded_parameters_features} (which gives $\|x\|_2\le L$), $\lVert x \rVert_{\Sigma_{t-1}} \leq \sqrt{\lambda_{\max}(\Sigma_{0})} \lVert x \rVert_{2} \leq L \sqrt{\lambda_{\max}(\Sigma_{0})}$.
By \cref{lem::wsb::ucb}, together with the fact that $\alpha_{t-1}^{\wsb} \lVert x \rVert_{\Sigma_{t-1}} \leq L \sqrt{\lambda_{\max}(\Sigma_{0})} \alpha_{t-1}^{\wsb}$, we have for any $\delta\in(0,1)$, with probability at least $1-\delta$, that
\begin{equation*}
\langle X_{t}^{*},\theta_{t}^{*}\rangle \leq \langle X_{t}^{*},\mu_{t-1}\rangle + L\sqrt{\lambda_{\max}(\Sigma_{0})} \alpha_{t-1}^{\wsb} + (\beta_{t-1}^{\wsb}(\delta/T) + \Pi_{t-1}) \lVert X_{t}^{*} \rVert_{\Sigma_{t-1}},
\end{equation*}
and with probability at least $1-\delta$ that
\begin{equation*}
\langle X_{t},\theta_{t}^{*}\rangle \geq \langle X_{t},\mu_{t-1}\rangle - L\sqrt{\lambda_{\max}(\Sigma_{0})} \alpha_{t-1}^{\wsb} - (\beta_{t-1}^{\wsb}(\delta/T) + \Pi_{t-1})\lVert X_{t} \rVert_{\Sigma_{t-1}}.
\end{equation*}
Hence, for any $\delta\in(0,1)$, we obtain, with probability at least $1-2\delta$, that
\begin{align*}
\langle X_{t}^{*}-X_{t}, \theta_{t}^{*}\rangle \leq & \langle X_{t}^{*}-X_{t}, \mu_{t-1}\rangle + 2 L\sqrt{\lambda_{\max}(\Sigma_{0})} \alpha_{t-1}^{\wsb} + (\beta_{t-1}^{\wsb}(\delta/T) + \Pi_{t-1}) \left( \lVert X_{t}^{*} \rVert_{\Sigma_{t-1}} + \lVert X_{t} \rVert_{\Sigma_{t-1}} \right)
\\ \leq & 2 L\sqrt{\lambda_{\max}(\Sigma_{0})} \alpha_{t-1}^{\wsb} + 2(\beta_{t-1}^{\wsb}(\delta/T)+\Pi_{t-1})\lVert X_{t} \rVert_{\Sigma_{t-1}},
\end{align*}
where the second inequality comes from the upper confidence bound selection criteria; $\langle X_{t}^{*},\mu_{t-1}\rangle + (\beta_{t-1}^{\wsb}(\delta/T) + \Pi_{t-1})\lVert X_{t}^{*} \rVert_{\Sigma_{t-1}} \leq \langle X_{t}, \mu_{t-1}\rangle + (\beta_{t-1}^{\wsb}(\delta/T) + \Pi_{t-1} )\lVert X_{t} \rVert_{\Sigma_{t-1}}$.
Thus, the regret $R_{T}$ can be upper bounded as follows:
\begin{equation*}
R_{T} \leq 2 L\sqrt{\lambda_{\max}(\Sigma_{0})} \sum_{t=1}^{T} \alpha_{t-1}^{\wsb} + 2\sum_{t=1}^{T} (\beta_{t-1}^{\wsb}(\delta/T) + \Pi_{t-1})\lVert X_{t} \rVert_{\Sigma_{t-1}},
\end{equation*}
where the first term corresponds to the bias component, while the second term accounts for the variance component. The variance term can be directly bounded using the fact that $\beta_{t}^{\wsb}(\delta)$ is non-decreasing in $t$ and $\Pi_{t}$ is non-increasing in $t$ (see its definition in \cref{lem::wsb::ucb}), along with the Cauchy-Schwarz inequality, \cref{lem::norm_bound,prop::determinant} with $p=1$:
\begin{align*}
& 2 \sum_{t=1}^{T} (\beta_{t-1}^{\wsb}(\delta/T) + \Pi_{t-1}) \lVert X_{t} \rVert_{\Sigma_{t-1}} 
\leq 2(\beta_{T}^{\wsb}(\delta/T)+\Pi_{0})\sum_{t=1}^{T} \lVert X_{t} \rVert_{\Sigma_{t-1}} 
\leq 2(\beta_{T}^{\wsb}(\delta/T)+\Pi_{0})\sqrt{T}\sqrt{\sum_{t=1}^{T} \lVert X_{t} \rVert^{2}_{\Sigma_{t-1}}}
\\ & \leq 2\sigma\sqrt{2\max\{1,L^{2} \lambda_{\max}(\Sigma_{0})/\sigma^2\}}(\beta_{T}^{\wsb}(\delta/T)+ \Pi_{0})\sqrt{T}\sqrt{d\sum_{t=1}^{T}\log\left(\frac{1}{w_{t-1,t}}\right) + \log\left(\frac{\det(\Sigma_{0})}{\det(\Sigma_{T})}\right)}
\\ & \leq 2\sigma\sqrt{2\max\{1,L^{2} \lambda_{\max}(\Sigma_{0})/\sigma^2\}}(\beta_{T}^{\wsb}(\delta/T)+ \Pi_{0})\sqrt{dT}\sqrt{\sum_{t=1}^{T}\log\left(\frac{1}{w_{t-1,t}}\right) + \log\left(1 + \frac{\Tr(\Sigma_{0})L^{2}\sum_{t=1}^{T}w_{t,T}}{d\sigma^{2}}\right)},
\end{align*}
which yields the desired bound:
\begin{equation*}
R_{T} \leq 2 L\sqrt{\lambda_{\max}(\Sigma_{0})} \sum_{t=1}^{T} \alpha_{t-1}^{\wsb} + 2^{3/2}\sigma\sqrt{\max\{1,L^{2} \lambda_{\max}(\Sigma_{0})/\sigma^2\}} (\beta_{T}^{\wsb}(\delta/T)+ \Pi_{0})\sqrt{dT\Lambda_{T}},
\end{equation*}
with $\Lambda_{T} = \sum_{t=1}^{T}\log\left(\frac{1}{w_{t-1,t}}\right) + \log\left(1 + \frac{\Tr(\Sigma_{0})L^{2}\sum_{t=1}^{T}w_{t,T}}{d\sigma^{2}}\right)$.
\end{proof}

\begin{proof}[Proof of \cref{cor::wsb::ucb::regret}]
By substituting $w_{s,t}=\gamma^{t-s}$ into the regret bound of $R_{T}$ (\cref{thm::wsb::ucb::regret}), we obtain the following expression:
\begin{align*}
R_{T} \leq 4 L^{2} \sqrt{d \lambda_{\max}(\Sigma_{0})} \frac{1}{(1-\gamma)^{3/2}} B_{T} + 2\sigma\sqrt{2\max\{1,L^{2} \lambda_{\max}(\Sigma_{0})/\sigma^2\}}(\beta_{T}^{\wsb}(\delta/T)+\Pi_{0})\sqrt{dT\Lambda_{T}'},
\end{align*}
with $\Lambda_{T} < \Lambda_{T}' = T\log\left(\frac{1}{\gamma}\right) + \log\left(1 + \frac{\Tr(\Sigma_{0})L^{2}}{d\sigma^{2}(1-\gamma)}\right)$, where we used that $\sum_{t=1}^{T}w_{t,T}=\sum_{t=1}^{T}\gamma^{T-t}=\frac{1-\gamma^{T}}{1-\gamma}<\frac{1}{1-\gamma}$ and that
\begin{align*}
\sum_{t=1}^{T} \sum_{k=1}^{t-1} \sqrt{\sum_{s=1}^{k} w_{s,t-1}} \lVert \theta_{k}^{*}-\theta_{k+1}^{*} \rVert_{2} & = \sum_{t=1}^{T} \sum_{k=1}^{t-1} \sqrt{\sum_{s=1}^{k} \gamma^{t-s-1}} \lVert \theta_{k}^{*}-\theta_{k+1}^{*} \rVert_{2} 
\\ & = \sum_{t=1}^{T} \sum_{k=1}^{t-1} \sqrt{\frac{\gamma^{t-1}(\gamma^{-k}-1)}{1-\gamma}} \lVert \theta_{k}^{*}-\theta_{k+1}^{*} \rVert_{2} 
\\ & =  \sum_{k=1}^{T-1} \sum_{t=k+1}^{T} \sqrt{\frac{\gamma^{t-1}(\gamma^{-k}-1)}{1-\gamma}} \lVert \theta_{k}^{*}-\theta_{k+1}^{*} \rVert_{2} 
\\ & = \sum_{k=1}^{T-1} \sum_{t=k+1}^{T} \sqrt{\gamma^{t-1}}\frac{\sqrt{\gamma^{-k}-1}}{\sqrt{1-\gamma}} \lVert \theta_{k}^{*}-\theta_{k+1}^{*} \rVert_{2} 
\\ & = \sum_{k=1}^{T-1} \frac{\sqrt{\gamma^{k}} - \sqrt{\gamma^{T}}}{1-\sqrt{\gamma}}\frac{\sqrt{\gamma^{-k}-1}}{\sqrt{1-\gamma}} \lVert \theta_{k}^{*}-\theta_{k+1}^{*} \rVert_{2} 
\\ & \leq \sum_{k=1}^{T-1} \frac{\sqrt{\gamma^{k}} - \sqrt{\gamma^{T}}}{(1-\sqrt{\gamma})(\frac{1+\sqrt{\gamma}}{2})}\frac{\sqrt{\gamma^{-k}-1}}{\sqrt{1-\gamma}} \lVert \theta_{k}^{*}-\theta_{k+1}^{*} \rVert_{2} 
\\ & = 2 \sum_{k=1}^{T-1} \frac{\sqrt{\gamma^{k}} - \sqrt{\gamma^{T}}}{1-\gamma} \frac{\sqrt{\gamma^{-k}-1}}{\sqrt{1-\gamma}} \lVert \theta_{k}^{*}-\theta_{k+1}^{*} \rVert_{2} 
\\ & = 2 \sum_{k=1}^{T-1} \frac{(\sqrt{\gamma^{k}} - \sqrt{\gamma^{T}})\sqrt{\gamma^{-k}-1}}{(1-\gamma)^{3/2}} \lVert \theta_{k}^{*}-\theta_{k+1}^{*} \rVert_{2} 
\\ & \leq 2 \sum_{k=1}^{T-1} \frac{\sqrt{\gamma^{k}} \sqrt{\gamma^{-k}}}{(1-\gamma)^{3/2}} \lVert \theta_{k}^{*}-\theta_{k+1}^{*} \rVert_{2} 
\\ & = 2 \sum_{k=1}^{T-1} \frac{1}{(1-\gamma)^{3/2}} \lVert \theta_{k}^{*}-\theta_{k+1}^{*} \rVert_{2} 
\\ & = \frac{2}{(1-\gamma)^{3/2}} B_{T}.
\end{align*}
Since the regret bound contains a term of the form $T\sqrt{\log(1/\gamma)}$, the discount factor $\gamma$ cannot become smaller than $1/T$, i.e., $\gamma\geq1/T$. Consequently, this leads to the inequality $\log(1/\gamma)\leq C (1-\gamma)$ where $C=\log(T)/(1-1/T)$.
By disregarding logarithmic dependencies on $T$, we can bound the regret as follows:
\begin{equation*}
R_{T} \leq \tilde{\mathcal{O}} \left( \frac{\sqrt{d \lambda_{\max}(\Sigma_{0})} B_{T}}{(1-\gamma)^{3/2}}  + dT\sqrt{1-\gamma} \right).
\end{equation*}
Setting $\lambda_{\max}(\Sigma_{0}) = 1/d$, this simplifies to
\begin{equation*}
R_{T} \leq \tilde{\mathcal{O}} \left( \frac{ B_{T}}{(1-\gamma)^{3/2}}  + dT\sqrt{1-\gamma} \right).
\end{equation*}
When $B_{T}$ is small, specifically $B_{T}<d/T$, choosing $\gamma=1-1/T$ yields the regret bound $R_{T} \leq \tilde{\mathcal{O}}(d\sqrt{T})$. For cases where $B_{T}$ is larger, i.e., $B_{T}\geq d/T$,  setting $\gamma = 1 - \sqrt{B_{T}/dT}$ results in $R_{T} \leq \tilde{\mathcal{O}}(B_{T}^{1/4}(dT)^{3/4})$, which completes the proof.
\end{proof}

\subsection{Randomized Exploration through WSB Perturbation} \label{sec::appendix:::wbs::random_exp}

Our analysis builds on \citet{kim2020randomized}, who introduced perturbation-based methods for non-stationary linear contextual bandits. But we adopt the refined weighted analysis of \citet{wang2023revisit}, which allows us to extend their results to the WSB framework and obtain improved regret bounds.

Let $\tilde{f}_{t}(x)$ denote the algorithm-specific selection criterion.  
For instance:
\begin{itemize}
    \item \texttt{WSB-RandLinUCB}: $\tilde{f}_{t}(x) = \langle x, \mu_{t-1}\rangle + \eta_t \lVert x \rVert_{\Sigma_{t-1}}$, with $\eta_{t} \sim \cN(0,a^{2})$ a scalar perturbation.
    \item \texttt{WSB-LinTS}: $\tilde{f}_{t}(x) = \langle x, \mu_{t-1}\rangle + x^{\top}\Sigma_{t-1}^{1/2}\eta_{t}$, with $\eta_{t} \sim \cN(0,a^{2}\I_d)$ a $d$-dimensional Gaussian vector. Equivalently, one may write $\tilde{f}_{t}(x) = \langle x, \mu_{t-1}\rangle + \eta_{t,x}\lVert x \rVert_{\Sigma_{t-1}}$ with $\eta_{t,x}\sim \cN(0,a^{2})$.
\end{itemize}

The central idea is that randomized algorithms can be analyzed by controlling the effect of (in this case) Gaussian perturbations. Concentration bounds ensure that perturbations remain bounded, while anti-concentration guarantees that the optimal action is not systematically discarded. Following \citet{kim2020randomized}, we define auxiliary events that combine these perturbation properties with our WSB confidence bounds (\cref{lem::wsb::ucb}). With these events in place, the refined analysis of \citet{wang2023revisit} applies directly, yielding improved regret guarantees for randomized exploration without the need for local norms.

We introduce the following events, which control both estimation error and random perturbations:
\begin{align*}
    \cE^{\text{WSB}} & \equiv \{ \forall t\geq1,\forall x\in\cX_{t} : \lvert \langle x, \mu_{t} - \bar{\mu}_{t} \rangle \rvert \leq c_{1} \lVert x \rVert_{\Sigma_{t-1}}\}, \\
    \cE^{\text{Conc.}}_{t} & \equiv  \{ \forall x\in\cX_{t} : \lvert \tilde{f}_{t}(x) - \langle x, \mu_{t}\rangle \rvert \leq c_{2} \lVert x \rVert_{\Sigma_{t-1}}\}, \\
    \cE^{\text{Anti-Conc.}}_{t} & \equiv  \{ \tilde{f}_{t}(X_{t}^{*}) - \langle X_{t}^{*}, \mu_{t} \rangle > c_{1} \lVert X_{t}^{*} \rVert_{\Sigma_{t-1}}\}.
\end{align*}

The next lemma adapts \citet[Theorem~3]{kim2020randomized} to the WSB framework.
\begin{lemma} \label{lem::wsb::regret::rand_exp}
Let $p_{1},p_{2},p_{3}\in(0,1)$,  a time horizon $T\in\N$, and suppose there exists constants $c_{1},c_{2}\geq1$ which may depend on $T$ such that $\P(\cE^{\text{WSB}}) \geq 1 - p_{1}$, $\P(\cE^{\text{Conc.}}_{t}) \geq 1 - p_{2}$, and $\P(\cE^{\text{Anti-Conc.}}_{t}) \geq p_{3}$.
Then, for any prior $\pi(\theta) = \mathcal{N}(\theta \vert \mu_{0},\Sigma_{0})$, the following inequality holds for all posteriors $\rho_{t}(\theta) = \cN(\theta \vert \mu_{t},\Sigma_{t})$ and $t \in [T]$ simultaneously:
\begin{equation*}
    \langle X_{t}^{*} - X_{t}, \bar{\mu}_{t} \rangle \leq p_{2} + (c_{1} + c_{2})\left( 1+\frac{2}{p_{3}-p_{2}} \right) \lVert X_{t} \rVert_{\Sigma_{t-1}},
\end{equation*}
where $X_{t}=\argmax_{x\in\cX_{t}}\tilde{f}_{t}(x)$ with $\tilde{f}_{t}(x)$ denoting the algorithm-specific selection criterion. 
\end{lemma}
\begin{proof}[Proof of \cref{lem::wsb::regret::rand_exp}]
The proof follows directly from analogous steps as in \citet[Theorem~3]{kim2020randomized}, with all arguments carrying over after replacing their WRLS confidence events and local norms with our WSB concentration events and the norm $\lVert \cdot \rVert_{\Sigma_{t-1}}$, and using our notation. 
\end{proof}

\begin{proof}[Proof of \cref{thm::wsb::regret::rand_exp}]
We begin with the standard decomposition of the regret:
\begin{equation*}
    R_{T} \leq 2L\sqrt{\lambda_{\max}(\Sigma_{0})}\sum_{t=1}^{T} \alpha_{t-1}^{\wsb} + \sum_{t=1}^{T} \langle X_{t}^{*} - X_{t}, \bar{\mu}_{t} \rangle.
\end{equation*}
Applying \cref{lem::wsb::regret::rand_exp} to the last term yields
\begin{align*}
   \sum_{t=1}^{T} \langle X_{t}^{*} - X_{t}, \bar{\mu}_{t} \rangle 
   = \sum_{t=1}^{T} \langle X_{t}^{*} - X_{t}, \bar{\mu}_{t} \rangle \,\mathbbm{1}_{\{\cE^{\text{WSB}}\}} + T \P(\overline{\cE^{\text{WSB}}})
   \leq (c_{1} + c_{2})\left( 1+\frac{2}{p_{3}-p_{2}} \right) \sum_{t=1}^{T} \lVert X_{t} \rVert_{\Sigma_{t-1}} + T(p_{1}+ p_{2}).
\end{align*}
By Cauchy–Schwarz inequality,
\begin{align*}
   \sum_{t=1}^{T} \lVert X_{t} \rVert_{\Sigma_{t-1}} 
   &\leq \sqrt{T} \sqrt{\sum_{t=1}^{T} \lVert X_{t} \rVert_{\Sigma_{t-1}}^{2}}
   \leq \sqrt{T}\sqrt{d\sum_{t=1}^{T}\log\left(\frac{1}{w_{t-1,t}}\right) + \log\left(\frac{\det(\Sigma_{0})}{\det(\Sigma_{T})}\right)} \\
   &\leq \sqrt{dT}\sqrt{\sum_{t=1}^{T}\log\left(\frac{1}{w_{t-1,t}}\right) + \log\left(1 + \frac{\Tr(\Sigma_{0})L^{2}\sum_{t=1}^{T}w_{t,T}}{d\sigma^{2}}\right)},
\end{align*}
where the last two inequalities follows from \cref{lem::norm_bound,prop::determinant} with $p=1$. 
Substituting this bound completes the proof.
\end{proof}

The crucial step in the above proof is the control of the drift-related term $2L\sqrt{\lambda_{\max}(\Sigma_{0})}\sum_{t=1}^{T} \alpha_{t-1}^{\wsb}$. This is precisely where the refined weighted analysis of \citet{wang2023revisit} enters, which avoids the virtual windowing arguments used in prior work and eliminates the need for \emph{local norms}. It is this refinement in the treatment of drift that yields the improved regret guarantees in our WSB analysis.

\subsubsection{Randomized Upper Confidence Bound (\texttt{WSB-RandLinUCB})} \label{sec::appendix:::wbs::randucb}

The following lemma, which follows from \citet[Lemmas~4, 5, and 6]{kim2020randomized}, establishes high-probability control of the events $\cE^{\text{WSB}}$, $\cE^{\text{Conc.}}_{t}$, and $\cE^{\text{Anti-Conc.}}_{t}$ for \texttt{WSB-RandLinUCB} with suitable choices of $c_1$ and $c_2$.

\begin{lemma} \label{lem::wsb::randucb::events}
For \texttt{WSB-RandLinUCB}, the arm-selection criterion is $\tilde{f}_{t}(x) = \langle x, \mu_{t} \rangle + \eta_{t}\lVert x \rVert_{\Sigma_{t-1}}$, with $\eta_{t} \sim \cN(0,a^2)$.
Then, the following hold with appropriate constant choices:
\begin{itemize}
    \item ($\cE^{\mathrm{WSB}}$). If $c_{1}=\sqrt{4\log(T)+d\log(1+\frac{\Tr(\Sigma_{0})L^{2}\sum_{t=1}^{T}w_{t,T}^{2}}{d\sigma^{2}})} + \Pi_{0}$, then $\P(\cE^{\mathrm{WSB}}) \geq 1-1/T $.
    \item ($\cE^{\mathrm{Conc.}}_{t}$). If $c_2 = a\sqrt{2\log(T/2)}$, then $\P(\overline{\cE^{\mathrm{Conc.}}_{t}}) \leq 1/T$.
    \item ($\cE^{\mathrm{Anti\text{-}Conc.}}_{t}$). If $a^2 = 14c_{1}^{2}$, then $\P(\cE^{\mathrm{Anti\text{-}Conc.}}_{t}) \geq \exp(-1/4)/(8\sqrt{\pi})$.
\end{itemize}
\end{lemma}

\begin{proof}[Proof of \cref{lem::wsb::randucb::events}]
The three statements follow directly from \citet[Lemmas~4, 5, and 6]{kim2020randomized}, respectively. The arguments extend directly to our WSB setting after replacing the WRLS confidence and \emph{local norms} with our WSB concentration and the norm $\lVert \cdot \rVert_{\Sigma_{t-1}}$. 
\end{proof}

\begin{proof}[Proof of \cref{cor::wsb::randucb::regret}]
The argument follows by analogous reasoning to the proof of \cref{cor::wsb::ucb::regret}. We have
\begin{equation*}
    R_{T} \leq 4 L^{2} \sqrt{d \lambda_{\max}(\Sigma_{0})} \frac{1}{(1-\gamma)^{3/2}} B_{T} 
    + (c_{1}+c_{2}) \left( 1 + \frac{2}{p_{3}-p_{2}} \right) \sqrt{dT\Lambda_{T}'} 
    + T(p_{1} + p_{2}),
\end{equation*}
where $\Lambda_{T}'$ is given as in the proof of \cref{cor::wsb::ucb::regret}.
With the choices of $c_{1}$ and $c_{2}$, this simplifies in $\tilde{\mathcal{O}}$-notation to
\begin{equation*}
    R_{T} \leq \tilde{\mathcal{O}}\left( 
    \frac{\sqrt{d \lambda_{\max}(\Sigma_{0})} B_{T}}{(1-\gamma)^{3/2}} 
    + dT\sqrt{1-\gamma} \right).
\end{equation*}
Setting $\lambda_{\max}(\Sigma_{0}) = 1/d$, we obtain
\begin{equation*}
    R_{T} \leq \tilde{\mathcal{O}}\!\left( 
    \frac{B_{T}}{(1-\gamma)^{3/2}} 
    + dT\sqrt{1-\gamma} \right).
\end{equation*}
If $B_{T}$ is small, specifically $B_{T}<d/T$, choosing $\gamma = 1 - 1/T$ yields $R_{T} \leq \tilde{\mathcal{O}}(d\sqrt{T})$.
On the other hand, if $B_{T} \geq d/T$, setting $\gamma = 1 - \sqrt{B_{T}/dT}$ gives $R_{T} \leq \tilde{\mathcal{O}}(B_{T}^{1/4}(dT)^{3/4})$. This completes the proof.
\end{proof}

\subsubsection{Thompson Sampling (\texttt{WSB-LinTS})} \label{sec::appendix:::wbs::ts}

The following lemma, which follows from \citet[Lemmas~4, 5, and 6]{kim2020randomized}, establishes high-probability control of the events $\cE^{\text{WSB}}$, $\cE^{\text{Conc.}}_{t}$, and $\cE^{\text{Anti-Conc.}}_{t}$ for \texttt{WSB-LinTS} with suitable choices of $c_1$ and $c_2$.

\begin{lemma} \label{lem::wsb::ts::events}
For \texttt{WSB-LinTS}, the arm-selection criterion is $\tilde{f}_{t}(x) = \langle x, \mu_{t} \rangle + x^\top \Sigma_{t-1}^{1/2}\eta_{t}$ with $\eta_t \sim \cN(0,a^2\I_d)$. Equivalently, $\tilde{f}_{t}(x) = \langle x, \mu_{t} \rangle + \eta_{t,x}\lVert x \rVert_{\Sigma_{t-1}}$, with $\eta_{t,x} \sim \cN(0,a^2)$. Then, the following hold with appropriate constant choices:
\begin{itemize}
    \item ($\cE^{\mathrm{WSB}}$). If $c_{1}=\sqrt{4\log(T)+d\log(1+\frac{\Tr(\Sigma_{0})L^{2}\sum_{t=1}^{T}w_{t,T}^{2}}{d\sigma^{2}})} + \Pi_{0}$, then $\P(\cE^{\mathrm{WSB}}) \geq 1-1/T $.
    \item ($\cE^{\mathrm{Conc.}}_{t}$). If $c_2 = a\sqrt{\log(KT/2)}$, then $\P(\overline{\cE^{\mathrm{Conc.}}_{t}}) \leq 1/T$.
    \item ($\cE^{\mathrm{Anti\text{-}Conc.}}_{t}$). If $a^2 = 14c_{1}^{2}$, then $\P(\cE^{\mathrm{Anti\text{-}Conc.}}_{t}) \geq \exp(-1/4)/(8\sqrt{\pi})$.
\end{itemize}
\end{lemma}

\begin{proof}[Proof of \cref{lem::wsb::ts::events}]
The three statements follow directly from \citet[Lemmas~4, 5, and 6]{kim2020randomized}, respectively. The arguments extend to our WSB setting after replacing the WRLS confidence and \emph{local norms} 
with the WSB concentration and the norm $\lVert \cdot \rVert_{\Sigma_{t-1}}$.
\end{proof}

\begin{proof}[Proof of \cref{cor::wsb::ts::regret}]
The argument follows by analogous reasoning to the proof of \cref{cor::wsb::ucb::regret,cor::wsb::randucb::regret};
\begin{equation*}
    R_{T} \leq 4 L^{2} \sqrt{d \lambda_{\max}(\Sigma_{0})} \frac{1}{(1-\gamma)^{3/2}} B_{T} 
    + (c_{1}+c_{2}) \left( 1 + \frac{2}{p_{3}-p_{2}} \right) \sqrt{dT\Lambda_{T}'} 
    + T(p_{1} + p_{2}),
\end{equation*}
where $\Lambda_{T}'$ is given as in the proof of \cref{cor::wsb::ucb::regret}. With the choices of $c_{1}$ and $c_{2}$, this simplifies in $\tilde{\mathcal{O}}$-notation to
\begin{equation*}
    R_{T} \leq \tilde{\mathcal{O}}\left( 
    \frac{\sqrt{d \lambda_{\max}(\Sigma_{0})} B_{T}}{(1-\gamma)^{3/2}} 
    + dT\sqrt{\log(K)}\sqrt{1-\gamma} \right).
\end{equation*}
Setting $\lambda_{\max}(\Sigma_{0}) = 1/d$, we obtain
\begin{equation*}
    R_{T} \leq \tilde{\mathcal{O}}\left( 
    \frac{B_{T}}{(1-\gamma)^{3/2}} 
    + dT\sqrt{\log(K)}\sqrt{1-\gamma} \right).
\end{equation*}
If $B_{T}$ is small, specifically $B_{T}<d/T$, choosing $\gamma = 1 - 1/T$ yields $R_{T} \leq \tilde{\mathcal{O}}(d\sqrt{\log(K)T})$. On the other hand, if $B_{T} \geq d/T$, setting $\gamma = 1 - \sqrt{B_{T}/dT\sqrt{\log(K)}}$ gives $R_{T} \leq \tilde{\mathcal{O}}(B_{T}^{1/4}\log(K)^{3/8}(dT)^{3/4})$. 
\end{proof}

\setcounter{equation}{0}
\renewcommand{\theequation}{\thesection.\arabic{equation}}
\setcounter{theorem}{0}
\renewcommand{\thetheorem}{\thesection.\arabic{theorem}}
\setcounter{lemma}{0}
\renewcommand{\thelemma}{\thesection.\arabic{lemma}}
\setcounter{proposition}{0}
\renewcommand{\theproposition}{\thesection.\arabic{proposition}}
\setcounter{corollary}{0}
\renewcommand{\thecorollary}{\thesection.\arabic{corollary}}
\setcounter{definition}{0}
\renewcommand{\thedefinition}{\thesection.\arabic{definition}}
\setcounter{assumption}{0}
\renewcommand{\theassumption}{\thesection.\arabic{assumption}}
\setcounter{example}{0}
\renewcommand{\theexample}{\thesection.\arabic{example}}
\setcounter{remark}{0}
\renewcommand{\theremark}{\thesection.\arabic{remark}}

\section{REGRET CURVES}
\label{sec::regret_curves}

\begin{figure}[h!]
\centering
\includegraphics[width=1.0\textwidth]{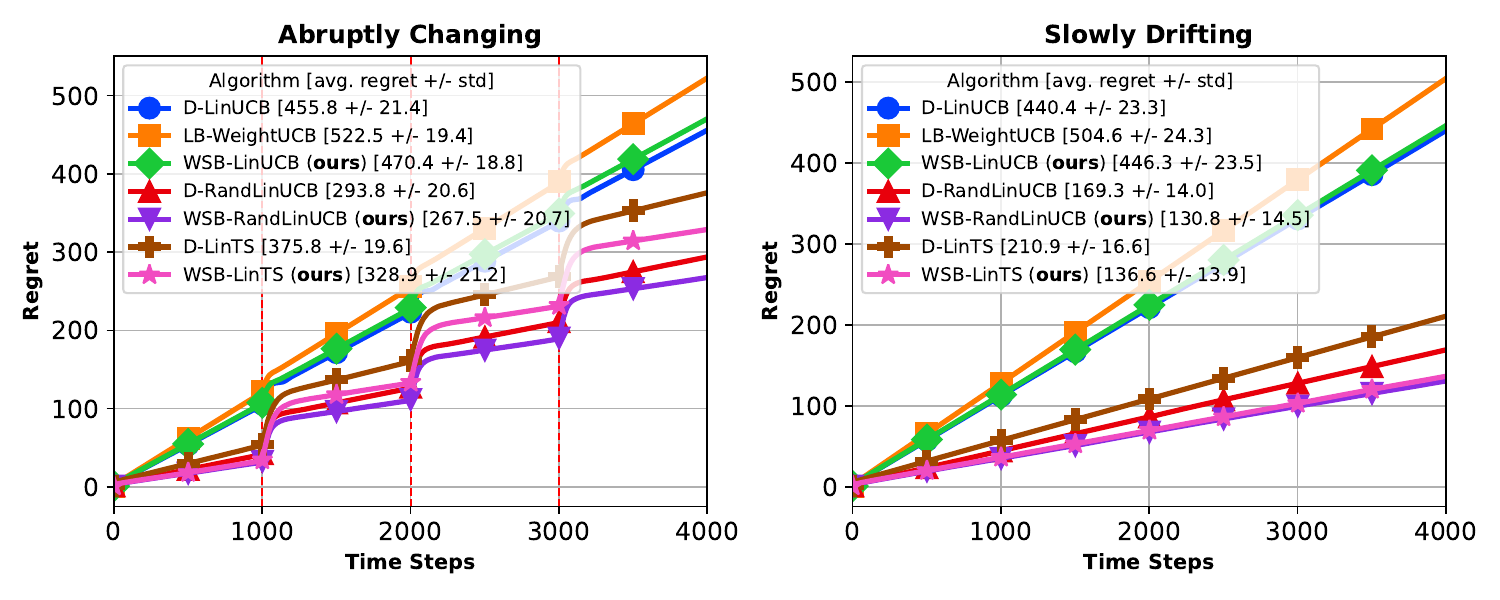}
\caption{Cumulative regret comparison with parameter dimension $d=2$ for the abruptly changing scenario (left column) and the slowly varying scenario (right column), averaged over $100$ independent trials. Vertical red dashed lines denote the environment change-points in the abruptly changing scenario, which remain unknown to the algorithms.} 
\label{fig::experiments_dim2}
\end{figure}
    
\begin{figure}[h!]
\centering
\includegraphics[width=1.0\textwidth]{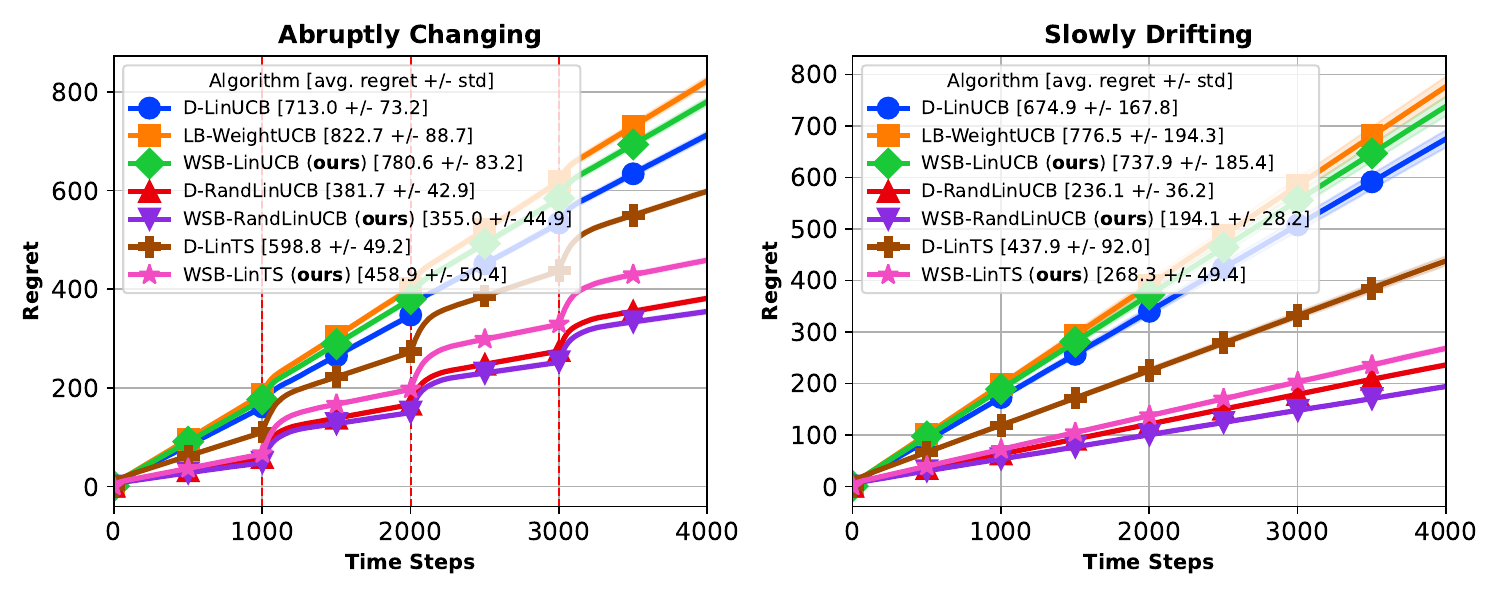}
\caption{Cumulative regret comparison with parameter dimension $d=4$ for the abruptly changing scenario (left column) and the slowly varying scenario (right column), averaged over $100$ independent trials. Vertical red dashed lines denote the environment change-points in the abruptly changing scenario, which remain unknown to the algorithms.} 
\label{fig::experiments_dim4}
\end{figure}
    
\begin{figure}[h!]
\centering
\includegraphics[width=1.0\textwidth]{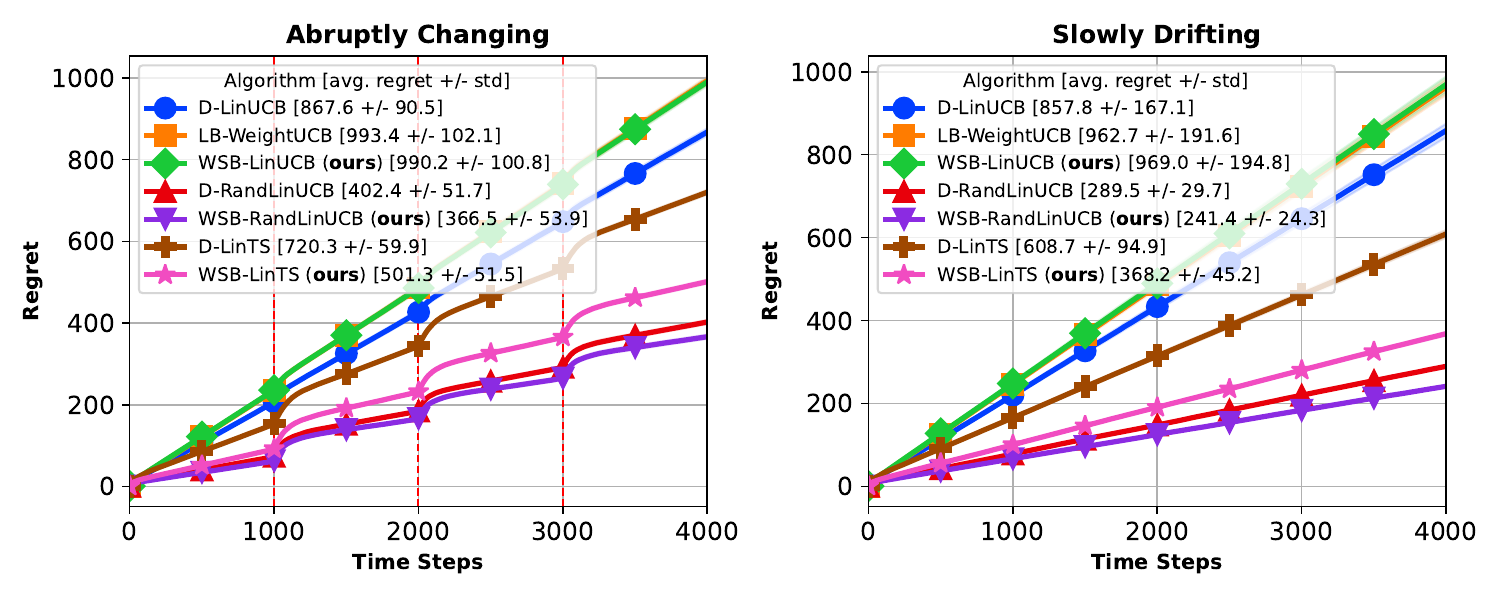}
\caption{Cumulative regret comparison with parameter dimension $d=6$ for the abruptly changing scenario (left column) and the slowly varying scenario (right column), averaged over $100$ independent trials. Vertical red dashed lines denote the environment change-points in the abruptly changing scenario, which remain unknown to the algorithms.} 
\label{fig::experiments_dim6}
\end{figure}

\begin{figure}[h!]
\centering
\includegraphics[width=1.0\textwidth]{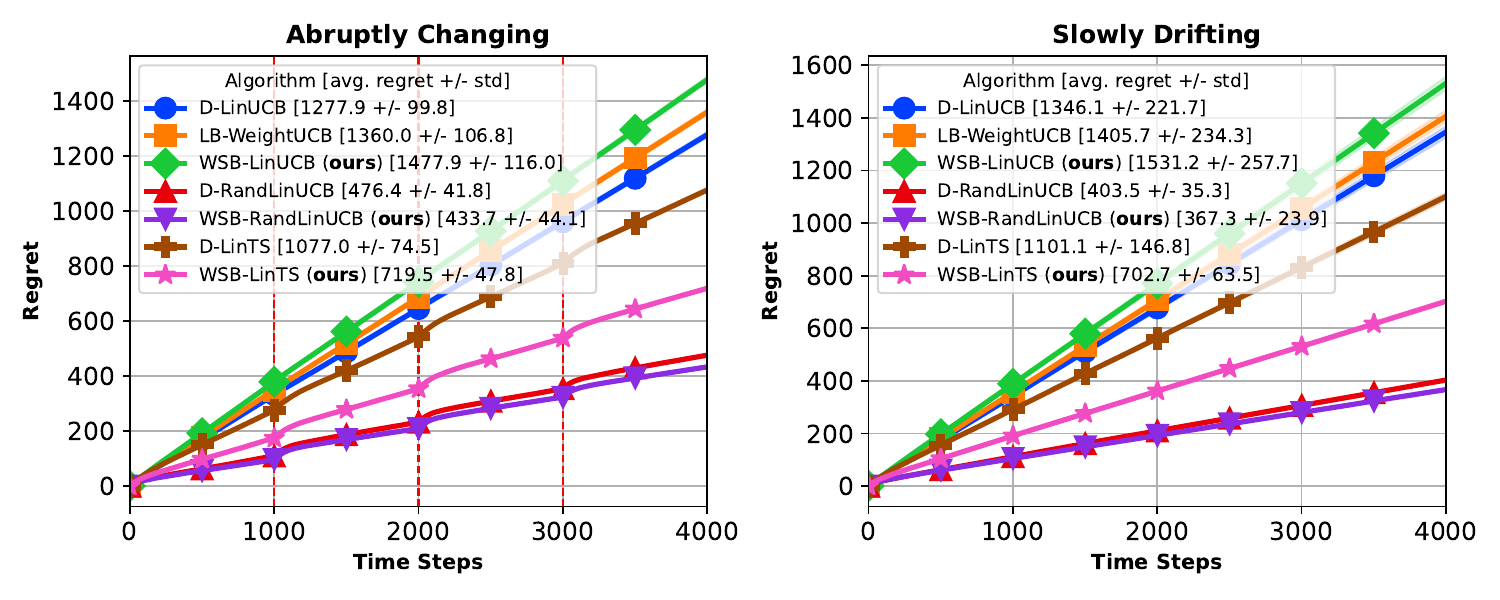}
\caption{Cumulative regret comparison with parameter dimension $d=16$ for the abruptly changing scenario (left column) and the slowly varying scenario (right column), averaged over $100$ independent trials. Vertical red dashed lines denote the environment change-points in the abruptly changing scenario, which remain unknown to the algorithms.} 
\label{fig::experiments_dim16}
\end{figure}
    
\begin{figure}[h!]
\centering
\includegraphics[width=1.0\textwidth]{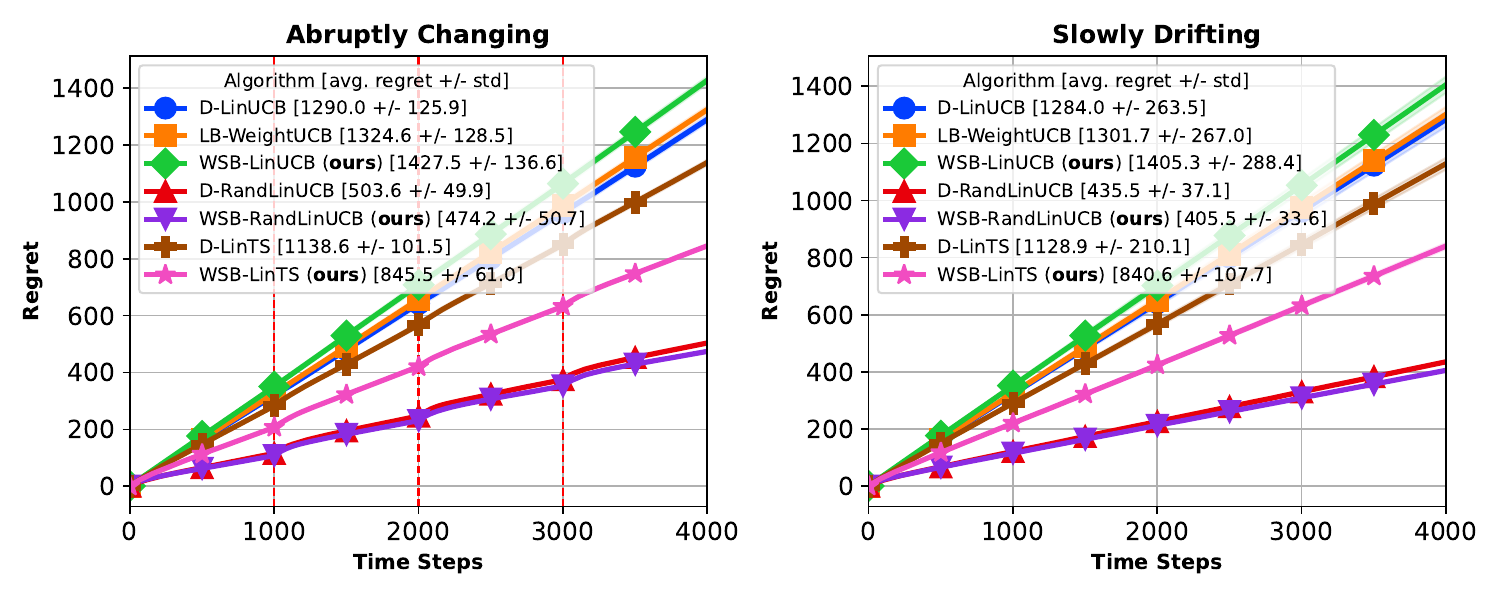}
\caption{Cumulative regret comparison with parameter dimension $d=32$ for the abruptly changing scenario (left column) and the slowly varying scenario (right column), averaged over $100$ independent trials. Vertical red dashed lines denote the environment change-points in the abruptly changing scenario, which remain unknown to the algorithms.} 
\label{fig::experiments_dim32}
\end{figure}

\clearpage
\section{ABLATION STUDY} \label{sec::ablation}

\begin{table}[h!]
\caption{Ablation study reporting cumulative regret (mean $\pm$ std) across context dimensions $d \in \{2,4,6,16,32\}$ in both abruptly changing and slowly drifting environments. The study evaluates sensitivity to prior misspecification, $\lVert \mu_0 \rVert_{2} \in \{1,10,100\}$, relative to the true parameter bound $S=1$.}
\label{tab:synthetic-regret-arms48-abrupt}
\begin{center}
\resizebox{\columnwidth}{!}{
\begin{tabular}{lccccc}
\multicolumn{6}{c}{\textbf{Abruptly Changing}} \\
\toprule
\textbf{ALGORITHM} & $d=2$ & $d=4$ & $d=6$ & $d=16$ & $d=32$ \\
\midrule
\texttt{WSB-LinUCB} ($\lVert\mu_0\rVert_{2}=1$)        & $ 474.5 \pm   17.9$ & $ 788.3 \pm   84.0$ & $ 996.6 \pm  107.3$ & $1480.6 \pm  115.4$ & $1429.9 \pm  135.6$ \\
\texttt{WSB-LinUCB} ($\lVert\mu_0\rVert_{2}=10$)       & $ 528.5 \pm   22.7$ & $ 878.5 \pm   91.9$ & $1092.2 \pm  113.7$ & $1536.1 \pm  121.8$ & $1446.7 \pm  137.4$ \\
\texttt{WSB-LinUCB} ($\lVert\mu_0\rVert_{2}=100$)      & $1642.6 \pm  408.9$ & $2228.5 \pm  401.9$ & $2258.4 \pm  329.5$ & $1928.6 \pm  170.4$ & $1516.2 \pm  140.9$ \\
\hdashline
\texttt{WSB-RandLinUCB} ($\lVert\mu_0\rVert_{2}=1$)    & $ 273.7 \pm   21.6$ & $ 352.6 \pm   44.2$ & $ 368.5 \pm   58.4$ & $ 443.5 \pm   45.7$ & $ 486.5 \pm   53.9$ \\
\texttt{WSB-RandLinUCB} ($\lVert\mu_0\rVert_{2}=10$)   & $ 627.2 \pm  134.5$ & $ 593.0 \pm  120.0$ & $ 562.7 \pm  101.1$ & $ 628.6 \pm  102.9$ & $ 606.2 \pm   99.4$ \\
\texttt{WSB-RandLinUCB} ($\lVert\mu_0\rVert_{2}=100$)  & $2515.1 \pm  781.0$ & $2399.7 \pm  592.5$ & $2066.5 \pm  432.2$ & $1733.9 \pm  195.3$ & $1321.3 \pm  170.4$ \\
\hdashline
\texttt{WSB-LinTS} ($\lVert\mu_0\rVert_{2}=1$)         & $ 332.2 \pm   19.8$ & $ 453.0 \pm   47.5$ & $ 505.2 \pm   50.6$ & $ 714.2 \pm   44.1$ & $ 851.5 \pm   60.2$ \\
\texttt{WSB-LinTS} ($\lVert\mu_0\rVert_{2}=10$)        & $ 472.6 \pm   68.4$ & $ 552.6 \pm   71.7$ & $ 582.6 \pm   58.6$ & $ 792.1 \pm   59.2$ & $ 892.7 \pm   80.9$ \\
\texttt{WSB-LinTS} ($\lVert\mu_0\rVert_{2}=100$)       & $2204.6 \pm  664.1$ & $2020.5 \pm  481.9$ & $1844.1 \pm  378.5$ & $1672.4 \pm  181.3$ & $1315.2 \pm  157.6$ \\
\bottomrule
\multicolumn{6}{c}{\textbf{Slowly Drifting}} \\
\toprule
\textbf{ALGORITHM} & $d=2$ & $d=4$ & $d=6$ & $d=16$ & $d=32$ \\
\midrule
\texttt{WSB-LinUCB} ($\lVert\mu_0\rVert_{2}=1$)        & $ 451.1 \pm   23.2$ & $ 743.6 \pm  186.9$ & $ 978.6 \pm  196.6$ & $1538.7 \pm  258.3$ & $1407.5 \pm  288.4$ \\
\texttt{WSB-LinUCB} ($\lVert\mu_0\rVert_{2}=10$)       & $ 520.1 \pm   40.8$ & $ 862.9 \pm  215.4$ & $1091.4 \pm  215.5$ & $1596.8 \pm  268.3$ & $1423.8 \pm  290.9$ \\
\texttt{WSB-LinUCB} ($\lVert\mu_0\rVert_{2}=100$)      & $1943.9 \pm  991.6$ & $2538.4 \pm  910.5$ & $2394.9 \pm  690.9$ & $1972.5 \pm  338.5$ & $1487.3 \pm  300.2$ \\
\hdashline
\texttt{WSB-RandLinUCB} ($\lVert\mu_0\rVert_{2}=1$)    & $ 138.7 \pm   16.8$ & $ 199.5 \pm   32.1$ & $ 247.8 \pm   28.3$ & $ 366.3 \pm   27.1$ & $ 407.4 \pm   34.2$ \\
\texttt{WSB-RandLinUCB} ($\lVert\mu_0\rVert_{2}=10$)   & $ 629.3 \pm  308.5$ & $ 539.9 \pm  205.9$ & $ 484.5 \pm  159.0$ & $ 569.3 \pm   99.7$ & $ 564.2 \pm  183.8$ \\
\texttt{WSB-RandLinUCB} ($\lVert\mu_0\rVert_{2}=100$)  & $2809.5 \pm 1742.7$ & $2671.7 \pm 1244.4$ & $2167.5 \pm  889.9$ & $1831.0 \pm  357.5$ & $1352.1 \pm  318.8$ \\
\hdashline
\texttt{WSB-LinTS} ($\lVert\mu_0\rVert_{2}=1$)         & $ 137.9 \pm   13.8$ & $ 268.6 \pm   48.7$ & $ 371.3 \pm   46.0$ & $ 706.2 \pm   63.0$ & $ 838.8 \pm  108.1$ \\
\texttt{WSB-LinTS} ($\lVert\mu_0\rVert_{2}=10$)        & $ 354.1 \pm  147.4$ & $ 413.6 \pm  100.0$ & $ 471.2 \pm   73.3$ & $ 791.1 \pm   81.4$ & $ 872.5 \pm  131.9$ \\
\texttt{WSB-LinTS} ($\lVert\mu_0\rVert_{2}=100$)       & $2371.1 \pm 1414.0$ & $2257.5 \pm  996.0$ & $1926.6 \pm  764.9$ & $1765.3 \pm  331.5$ & $1330.4 \pm  314.6$ \\
\bottomrule
\end{tabular}}
\end{center}
\end{table}

We evaluate empirical robustness by testing the sensitivity of the WSB-based algorithms to prior misspecification. Specifically, we vary the norm of the prior mean as $\lVert \mu_0 \rVert_2 \in \{1,10,100\}$, while the true parameter sequence satisfies the bound $S=1$. The results show that WSB is reasonably stable when the prior scale is comparable to, or moderately larger than, the true parameter scale. Moving from $\lVert \mu_0 \rVert_2=1$ to $\lVert \mu_0 \rVert_2=10$ increases regret, but the algorithms generally retain the same qualitative behavior across both abruptly changing and slowly drifting environments. In contrast, the extreme setting $\lVert \mu_0 \rVert_2=100$ leads to a substantial degradation in performance, especially for the randomized exploration methods. This confirms that the prior influence captured by the dynamic term $\Pi_t$ is practically relevant: although sequential posterior updates can reduce the effect of the prior over time, a prior mean far outside the true parameter scale can dominate early decisions and delay adaptation. Overall, the ablation highlights both the robustness and the limitations of WSB under prior misspecification.

\end{document}